\documentclass{article}

\usepackage[final]{corl_2025} 

\usepackage{amsmath}
\usepackage{amsthm}
\usepackage{amsfonts}
\usepackage{booktabs}
\usepackage{tcolorbox}
\usepackage[utf8]{inputenc} 
\usepackage[T1]{fontenc}    
\usepackage{url}                
\usepackage{booktabs}       
\usepackage{amsfonts}       
\usepackage{nicefrac}       
\usepackage{microtype}      
\usepackage{xcolor}         
\usepackage{float}
\usepackage{placeins}   
\usepackage{algpseudocode}  
\makeatletter
\algrenewcommand\algorithmicrequire{\textbf{Input:}}
\algrenewcommand\algorithmicensure {\textbf{Output:}}
\makeatother
\usepackage{multicol}
\usepackage{lipsum}
\usepackage{enumitem}
\usepackage{color, colortbl}
\definecolor{Gray}{gray}{0.95}
\definecolor{Highlight}{cmyk}{1,0.19,0,.22}
\usepackage{textgreek}
\usepackage{multirow}
\usepackage{ifthen}
\usepackage{graphicx}
\usepackage{caption}
\usepackage{algorithm}
\newfloat{algorithm}{t}{lop}
\usepackage{textcomp}
\usepackage{subcaption}
\usepackage{amsmath,amssymb, amsthm}
\usepackage{bbm}
\usepackage{wrapfig}
\definecolor{darkgreen}{rgb}{0.0, 0.4, 0.0}
\definecolor{darkgray}{gray}{0.3}
\definecolor{lightgray}{gray}{0.95}
\definecolor{highlight}{rgb}{1,0,0}
\usepackage{tcolorbox}
\usepackage{xcolor}
\definecolor{primary}{HTML}{003366}
\definecolor{accent}{HTML}{D62828}
\definecolor{bgbox}{HTML}{F8F9FA}
\definecolor{textcolor}{HTML}{222222}
\newcommand{\M}{\mathcal{M}}
\renewcommand{\S}{\mathcal{S}}
\newcommand{\A}{\mathcal{A}}
\newcommand{\OO}{\mathcal{O}}
\newcommand{\R}{\mathcal{R}}
\renewcommand{\P}{\mathcal{P}}

\newcommand{\Method}{ATK}

{}

\title{ATK: Automatic Task-driven Keypoint Selection for Robust Policy Learning}

\author{%
  \begin{tabular}{c}
    \vspace{0.1cm}
    Yunchu Zhang \quad
    Shubham Mittal \quad
    Zhengyu Zhang \quad
    Liyiming Ke \\[2pt]
    Siddhartha Srinivasa \quad
    Abhishek Gupta\\[2pt]
    \\Paul G. Allen School of Computer Science and Engineering
    \\University of Washington
    \end{tabular}
}

\begin{document}
\maketitle

\begin{abstract}
Visuomotor policies often suffer from perceptual challenges, where visual differences between training and evaluation environments degrade policy performance. Policies relying on state estimations, like 6D pose, require task-specific tracking and are difficult to scale, while raw sensor-based policies may lack robustness to small visual disturbances. In this work, we leverage 2D keypoints — spatially consistent features in the image frame — as a flexible state representation for robust policy learning and apply it to both sim-to-real transfer and real-world imitation learning. However, the choice of which keypoints to use can vary across objects and tasks. We propose a novel method, $\Method$, to automatically select keypoints in a task-driven manner so that the chosen keypoints are predictive of optimal behavior for the given task. Our proposal optimizes for a minimal set of keypoints that focus on task-relevant parts while preserving policy performance and robustness. We distill expert data (either from an expert policy in simulation or a human expert) into a policy that operates on RGB images while tracking the selected keypoints. By leveraging pre-trained visual modules, our system effectively encodes states and transfers policies to the real-world evaluation scenario  despite wide scene variations and perceptual challenges such as transparent objects, fine-grained tasks, and deformable objects manipulation. We validate $\Method$ on various robotic tasks, demonstrating that these minimal keypoint representations significantly improve robustness to visual disturbances and environmental variations. See all experiments and more details on our \href{https://yunchuzhang.github.io/ATK/}{\color{blue}{website}}.

\end{abstract}


\section{Introduction}
\label{sec:intro}

Though powerful in principle, visuomotor policy learning in practice often requires a significant number of samples to learn robust, generalizable policies ~\cite{pi0, kim2024openvla, guidedpolicy, levinehandeye, chi23diffusion}. To make this paradigm more practical, many methods use \textit{pretrained visual representations} ~\cite{dinov2, parisi22vision, nair22r3m}; these representations, often obtained through self-supervised learning objectives (such as reconstruction~\cite{he22mae}, future prediction~\cite{oordcpc} or contrastive learning~\cite{dinov2, chen20simclr, byollatent}), improve sample efficiency and robustness in many domains. However, despite this pretraining, the resulting policies can remain \textit{brittle}, i.e.,  responsive to distractors, object changes, and lighting changes, making them difficult to broadly deploy~\cite{aniredteaming, byovla}. This problem raises the question that motivates our research: \textit{How can we design general-purpose yet tailorable  representations of visual input that make policies robust to environmental variations but still transferable across scenarios?} 


In this work, we propose the use of \emph{keypoints} 
—a set of 2D pixel points in RGB images that can be tracked over time—as general-purpose visual state representations for robotic manipulation policies. Extensively used in computer vision~\cite{doersch23tapir,cotracker}, keypoints track specific semantically meaningful points on an image and have demonstrated robustness even in the presence of occlusion, lighting changes, and scale variations. Unlike pose-based methods, they do not rely on rigid structures, making them more suitable for tracking articulated and 
deformable objects. Additionally, keypoints naturally generalize across extreme object appearances, such as transparent, reflective, or fine-grained. Recent advances in keypoint tracking, driven by models trained on large-scale web data \cite{doersch23tapir, tang23dift}, reveal that keypoint tracking is surprisingly robust across diverse visual domains. These advantages support our contention that, when chosen appropriately, keypoints are promising candidates for powerful, robust visual representations.

\begin{figure}[t]
 \centering
 \includegraphics[width=\linewidth]{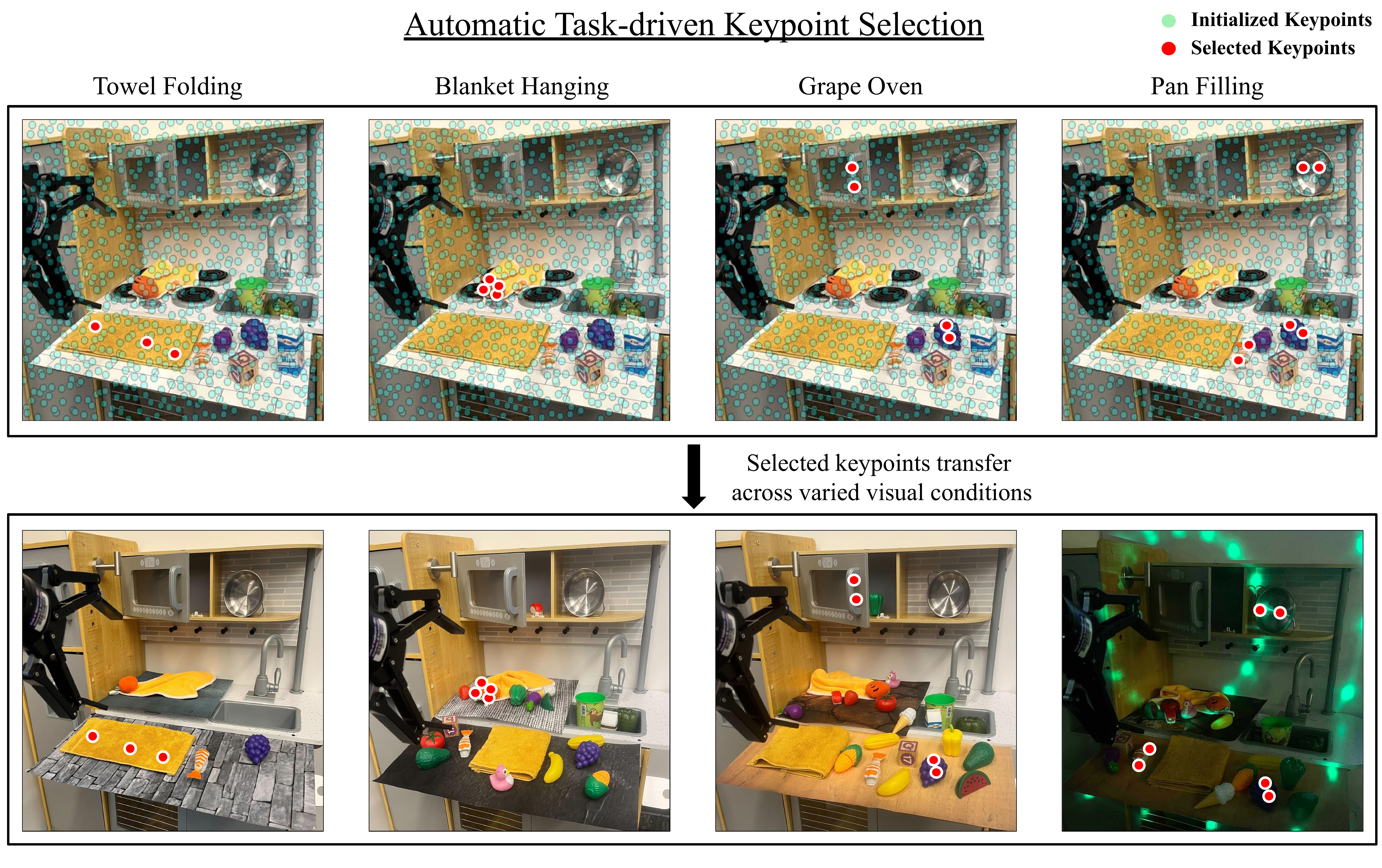}
 \caption{
 \textbf{(Top:)Automatically selected keypoint representations for different tasks in the same scene. }Scene representations vary depending on the desired functionality. \textbf{(Bottom:)} Robust generalization of policies learned keypoint representations to various positions, backgrounds, distractors, and lighting changes.}
 \label{fig:concept}
\end{figure}

Having established the value of keypoints, we next ask: \textit{What is the minimal set of task-relevant keypoints that can serve as an effective state representation for decision making?} Simply using all keypoints in a scene leads to inefficient redundancy, increases computational burden, and complicates tracking due to occlusion and point interference. Random sampling of points or selecting too few points risks overlooking critical task-relevant information, making the optimal policy unrealizable in the representation class. Crucially, the ideal set of keypoints varies from task to task, as shown in Fig~\ref{fig:concept}. Each task requires focusing on different parts of the scene, indicating that \textit{the minimal set of keypoints must be inherently task driven}. 


Our key insight, then, is that task objectives should inform both the selection of compact keypoint representations and the training of an optimal policy. 
We infer that a suitable task-driven representation is the \emph{minimal} set of keypoints that can sufficiently predict the optimal policy. Intuitively, this suggests that points that are not predictive of optimal actions can be dropped since they are not necessary for decision making. However, such reasoning creates a chicken-and-egg problem: a robust state representation is needed to learn the optimal policy, yet optimal policies are necessary to identify the appropriate state representation. Fortunately, when synthesizing policies via common data-driven learning methods, e.g., imitation learning ~\cite{zare2024survey} or student-teacher distillation of policies for sim-to-real transfer~\cite{rmaashish, chen21hand}, optimal actions are available to inform the choice of minimal, task-relevant keypoints. We propose a distillation-based algorithm, $\Method$, that uses a masking architecture to jointly select a minimal set of task-relevant keypoints and train a keypoint conditional policy via supervised learning. This minimal, task-specific keypoint representation retains the necessary task-relevant information, making the resulting policies naturally robust to environmental variations and transferable across scenarios with considerable visual differences, e.g., sim-to-real transfer. 


In sum, this work contributes: 
\begin{itemize}
\item A methodology for jointly \textbf{selecting a minimal set of task-relevant keypoints} and learning a policy conditioned on these keypoints. 
\item An empirical validation of this approach across a variety of real-world robot manipulation tasks in the \textbf{sim-to-real} setting, demonstrating robust sim-to-real transfer in settings with considerable visual variety. 
\item A demonstration of the efficacy of the proposed methodology in the \textbf{imitation learning} setting, with the resulting policies showing strong visual generalization while retaining high-precision dexterity.
\end{itemize}

\section{Problem Formulation}
\label{sec:prelim}
We study decision making in finite-horizon Markov Decision Processes (MDPs) defined by the tuple $\M = (\S, \OO, \A, \P, \rho_0, \R, \gamma)$, where $\S$ represents the Lagrangian state space (the compact, physical state of the system), $\OO$ is the observation space, $\A$ is the action space, $\P(s'|s, a)$ defines the transition dynamics, $\rho_0$ is the initial state distribution, $\R$ is the reward function, and $\gamma$ is the discount factor. In simulation, agents have access to the Lagrangian state $\S$, which provides a compact, complete description of the environment (e.g., object positions, velocities, etc.). In the real world, agents can access only sensor observations $\OO$ (e.g., RGB images). Although the real world might be partially observable, we assume that the current observation $o \in \OO$ is sufficient to make optimal decisions. The observation $o$ is produced by an invertible emission function $f$, such that $o = f(s)$. Our goal is to derive a visuomotor policy $\pi_\theta$ that is near optimal in the real world when acting on observations $o_t$.


\textbf{Simulation to Reality Transfer.} We aim to derive policies that operate from perceptual inputs for transfer from simulation to reality. Though we have ``privileged" information $s_t$ in simulation that can enable rapid learning of (near) expert policies $\pi^*(\cdot | s_t)$ via standard decision-making algorithms (including imitation learning, reinforcement learning, trajectory optimization, or motion planning), transferring these \emph{perception-based} policies $\pi_\theta(\cdot | o_t)$ from simulation to the real world operation is difficult. A key challenge is the perceptual gap between simulation and the real world.  This can be formalized using two MDPs: $\M_{\text{sim}} = (\OO_{\text{sim}}, \S, \A, \P, \rho_0, \R, \gamma)$ for simulation, and $\M_{\text{real}} = (\OO_{\text{real}}, \A, \P, \rho_0, \R, \gamma)$ for the real world. The same underlying state $s$ goes through different emission functions and leads to different observations (such as RGB images) in simulation, $o_{\text{sim}} = f_{\text{sim}}(s)$, vs the real world, $o_{\text{real}} = f_{\text{real}}(s)$. The challenge in transferring end-to-end visuomotor policies $\pi^*(a_t | o_t)$ from simulation to the real world lies in the mismatch between $\OO_{\text{sim}}$ and $\OO_{\text{real}}$.
 
\textbf{Imitation Learning.} The imitation learning setting is provided with an offline dataset of expert behavior data in lieu of a simulator. The dataset $\mathcal{D} = \{(o_i, a_i)\}_{i=1}^M$ is drawn from an expert -- $\pi^*$. Imitation learning settings both train and test in the real world from raw camera observations. However, for imitation learning to be visually ``robust," we consider training on a setting with one emission function, $f_{\text{train}}(s)$, and evaluating with another emission function, $f_{\text{test}}(s)$. We can formulate this via a simple dual MDP formulations: $\M_{\text{train}} = (\OO_{\text{train}}, \S, \A, \P, \rho_0, \R, \gamma)$ for training and $\M_{\text{test}} = (\OO_{\text{real}}, \A, \P, \rho_0, \R, \gamma)$ for evaluation. The same underlying state $s$ goes through different emission functions and leads to different observations at training, $o_{\text{train}} = f_{\text{train}}(s)$, and testing, $o_{\text{test}} = f_{\text{test}}(s)$. For instance, this approach could involve training for robotic manipulation on a clear tabletop and then testing with changes in background, lighting or distractor objects. The challenge lies in the mismatch between $\OO_{\text{train}}$ and $\OO_{\text{test}}$.

To address both challenges, we select a state representation that retains invariance between simulation and real-world $g_\text{sim}(o_{\text{sim}}) = g_\text{real}(o_{\text{real}})$, or train and test observations  $g_\text{train}(o_{\text{train}}) = g_\text{test}(o_{\text{test}})$. Though many such choices are feasible, this work focuses on keypoint-based representations.

\section{Task-Driven Automatic Keypoint Selection for Robust Policy Learning}

We aim to provide an input representation that can enable policy generalization and robustness for the transfer settings mentioned in Section~\ref{sec:prelim}. To this end, we propose the use of \textbf{2D keypoints} as the perceptual representation for sim-to-real transfer (Sec.~\ref{sec:keypoints}). The crux of our proposal lies in transferring only task-relevant parts of the observation by automatically \emph{selecting} a set of task-relevant keypoints. We propose an algorithm that integrates keypoint selection with policy training using a distillation process that relies on expert data to propose keypoints and obtain corresponding keypoint-based policies (Sec.~\ref{sec:distill}). Finally, we explain how to deploy these chosen keypoints and their trained policies in real-world evaluations (Sec.~\ref{sec:transfer}). 

\subsection{\textbf{Keypoints as Policy Representations}\label{sec:keypoints}} 

Keypoints, a widely used representation in computer vision, are salient locations in an image that are useful for identifying and describing objects or features. Formally, a keypoint is defined as a specific position \( k_i^t = (x_i^t, y_i^t) \) in the 2D image plane at time $t$. A set of $N$ keypoints,  \( \{k_i^t\}_{i=1}^N = (x_1^t, y_1^t,...,x_N^t, y_N^t) \), provides a compact scene representation. The number and selection of keypoints can be dynamically adjusted based on task complexity and requirements. However, what makes keypoints impactful for robust policy learning is the proliferation of robust tracking algorithms ~\cite{cotracker, doersch23tapir}, trained on web-scale data, that maintain dense correspondences across frames despite visual scene-level and instance-level variations. We bring this robustness to bear on policy learning. To track keypoints over time, we initialize keypoints $\{k_i^t\}_{i=1}^N$ at $t=0$ and then use tracking methods~\cite{yang23trackanything, cotracker, doersch23tapir} to maintain robust semantic correspondences of these points across time steps. We formalize tracking as a correspondence function \( h_{\mathcal{C}} \) that updates keypoint locations at each time step while providing correspondence measurement scores.

Given a set of initial keypoint positions,  \( \{k_i^t\}_{i=1}^N = (x_1^t, y_1^t,...,x_N^t, y_N^t) \), a particular set of their positions at time \( t \) is updated as: $\{k_i^t\}_{i=1}^N = h_{\mathcal{C}}(\{k_{i}^{t-1}\}_{i=1}^N, I_{t})$, where $I_{t}$ is the current image. The observation at time t, $o_t$, is an (ordered) set of $N$ keypoints $\{k_i^t\}_{i=1}^N$ that is updated iteratively through the correspondence function $h_{\mathcal{C}}$. The keypoint locations \( k_i^t = (x_i^t, y_i^t) \) correspond to the \emph{current} planar positions of points that semantically correspond to the initially chosen points $\{k_i^0\}_{i=1}^N$.

The core challenges in leveraging keypoints as a policy representation are the selection of initial keypoints $\{k_i^0\}_{i=1}^N$,  the tracking of them through time, and the transferring of resulting keypoint-based policies across widely varying deployment scenarios. The choice of keypoints is inherently task-specific since different tasks require focusing on distinct elements in the scene. For example, in the kitchen scene shown in Fig~\ref{fig:concept}, the keypoints on the blanket are crucial for the blanket-hanging task, whereas the pan-placement and grasping tasks require keypoints on both the pan and other objects.

\subsection{\textbf{Automatic Task-Driven Keypoint Selection: Training}}
\label{sec:distill}

Our work selects a \emph{minimal} set of task-relevant keypoints as a representation to enable robust policy transfer. Formally, we aim to identify a minimal set of \(N\) task-relevant keypoints, \(\{k_i\}_{i=1}^N\), that enable training a near-optimal policy while being easily tracked with \(h_{\mathcal{C}}\). Our keypoint selection is based on two criteria:  (1) \textbf{realizability of the optimal policy}, i.e., the selected keypoints must capture all necessary information to learn a near optimal policy for the task, and (2) \textbf{trackability}, i.e., the chosen keypoints must be reliably and consistently trackable using an available correspondence function $h_{\mathcal{C}}$. Realizability and trackability are naturally interconnected concepts; representations that are not trackable are not consistent, making it impossible to realize an optimal policy. \textit{Given the full set of $N$ candidate keypoints $\{k_i^t\}_{i=1}^N$, how do we select a minimal set of $K$ keypoints, with $K \ll N$, to satisfy realizability} \textit{and trackability?}

\begin{figure*}[h!]
    \centering
    \includegraphics[width=\textwidth]{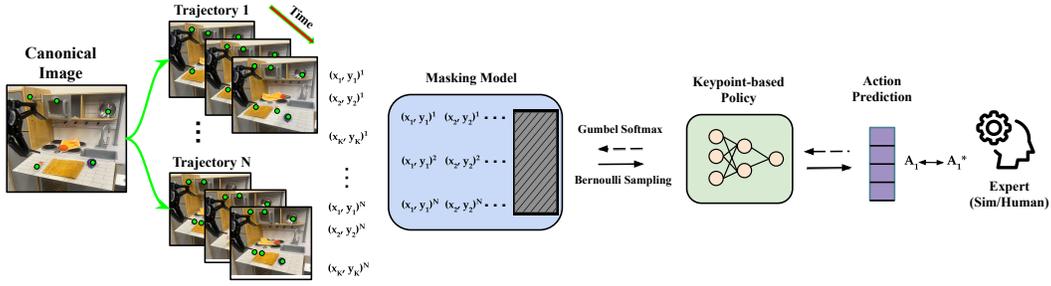}
    \caption{\footnotesize{\textbf{ATK automatically selects minimal yet necessary information for task execution} by distilling expert data (either from an expert policy in simulation or a human expert) into a policy that operates on a selective subset of keypoints and optimizing the selection mask. Once the keypoints are identified, they are transferred from the training set to the real-world evaluation scenario. Finally, the keypoint-based policy is transferred to the evaluation scenario, taking as input RGB images while tracking the transferred keypoints.}}
    \label{fig:kp_main}
\end{figure*}

Our insight is that the key workhorse in both imitation learning and sim-to-real distillation---\textit{supervised action prediction}---can directly inform both keypoint selection and subsequent policy learning. Given a dataset of expert observation-action tuples \(
\mathcal{D} = \{(o_i, a_i)\}_{i=1}^M
\), we typically learn a visuomotor policy via supervised learning: $\theta^* \;=\; \arg\min_{\theta}\; \frac{1}{M}\sum_{i=1}^M \bigl\lVert \pi_\theta(o_i) - a_i \bigr\rVert^2$. This can be easily generalized to more expressive maximum likelihood objectives~\cite{chi23diffusion, pi0}. Supervised learning objectives underpin both imitation learning and sim-to-real distillation. For imitation learning, the expert dataset comes from a human expert; in sim-to-real distillation, it comes from a privileged expert in simulation. In this work, we propose a \emph{masking}\textit{-based keypoint selection mechanism} that uses gradients from supervised learning to select a minimal set of task-relevant keypoints. 


To ensure that each keypoint remains semantically and spatially consistent across trajectories, we first identify a single image frame that captures the complete task context for solving the task, denoted as the \textit{canonical template} $I_{\text{can}}$. We then randomly sample \(C\) candidate keypoints 
\(\{k_j^0\}_{j=1}^C\) on the canonical image \(I_{\mathrm{can}}\). Given an expert dataset
\(\mathcal{D} = \bigl\{(o_i, a_i)\bigr\}_{i=1}^M,\) we use the correspondence function \(h_{\mathcal{C}}\) to align and propagate these candidates across all trajectories and timesteps, producing a keypoint-annotated dataset
\(\mathcal{D}_k = \bigl\{(o_{i,t},\,K_{i,t},\,a_i)\bigr\}_{i=1,t=1}^{M,T_i},\)
where \(K_{i,t}\) is the set of \emph{all} tracked keypoint positions for observation \(o_{i,t}\). From \(\mathcal{D}_k\), we jointly learn a sparse masking model \(\mathbb{M}_{\phi}\) and a downstream keypoint-based policy \(\pi^k_{\theta}\). As shown in Fig.~\ref{fig:kp_main}, the \(M\) candidate keypoints are fed into \(\mathbb{M}_{\phi}\), which outputs \(M\) independent Bernoulli probabilities, with each representing the likelihood of retaining the corresponding keypoint from the input candidate set. Sampling a binary mask \(m\in\{0,1\}^M\) then zeroes out the unselected keypoints, yielding a reduced set \(\widetilde K\). Finally, \(\widetilde K\) is passed to \(\pi^k_{\theta}\), which produces the action distribution, as is standard in imitation learning. This unified architecture first applies pointwise masking to select the minimal keypoint subset and then predicts actions from those selected points.


To select the \emph{minimal} set of $K$ keypoints, we enforce a sparse information bottleneck on the mask: 

\[
\min -\mathbb{E}_{(k,a) \sim \mathcal{D}} \log \pi_{\theta}^{k}\!\left(
    a_t^{*} \,\middle|\, 
    \mathbb{M}_{\phi}\!\bigl(\{k_i^{t}\}_{i=1}^{N}\bigr)
\right) + \alpha \| \mathbb{M}_{\phi}(\{k_i^t\}_{i=1}^N) \|_1.
\]


Intuitively, this training procedure filters out points that are (1) irrelevant to predicting optimal actions and (2) challenging to track using $h_{\mathcal{C}}$ since their representations over time $(\{k_i^t\}_{i=1}^N)$ are unreliable markers of optimal actions. Since the mask sampling is discrete and non-differentiable, we employ the Gumbel-softmax relaxation~\cite{gumbelsoftmax} to enable gradient-based optimization.


\subsection{\textbf{Inference with Task-Driven Keypoints}}
\label{sec:transfer}

The preceding procedure learns a masking model $\mathbb{M}_{\phi}$ and a policy $ \pi_{\theta}^{k}\!\left(
    a_t^{*} \,\middle|\, 
    \mathbb{M}_{\phi}\!\bigl(\{k_i^{t}\}_{i=1}^{N}\bigr)
\right)$. How can we use them for robust policy inference at test time? Whether it be a transfer from simulation to reality or from one imitation learning scenario at training time to another at test time, the inference procedure remains the same. For inference, it is important to transfer the minimal set of keypoints selected at training time to a variety of visually diverse test time scenarios. Since tracking of keypoints is performed by robust web-scale visual trackers \cite{cotracker, doersch23tapir}, once the \emph{initial} selected set of keypoints is identified in the real world at test time, subsequent tracking is not affected by the visual gap. This lets us focus solely on transferring the \emph{initial} set of keypoints.
\begin{wrapfigure}{r}{0.35\linewidth}
    \centering
    \vspace{-0.5cm}
    \includegraphics[width=\linewidth]{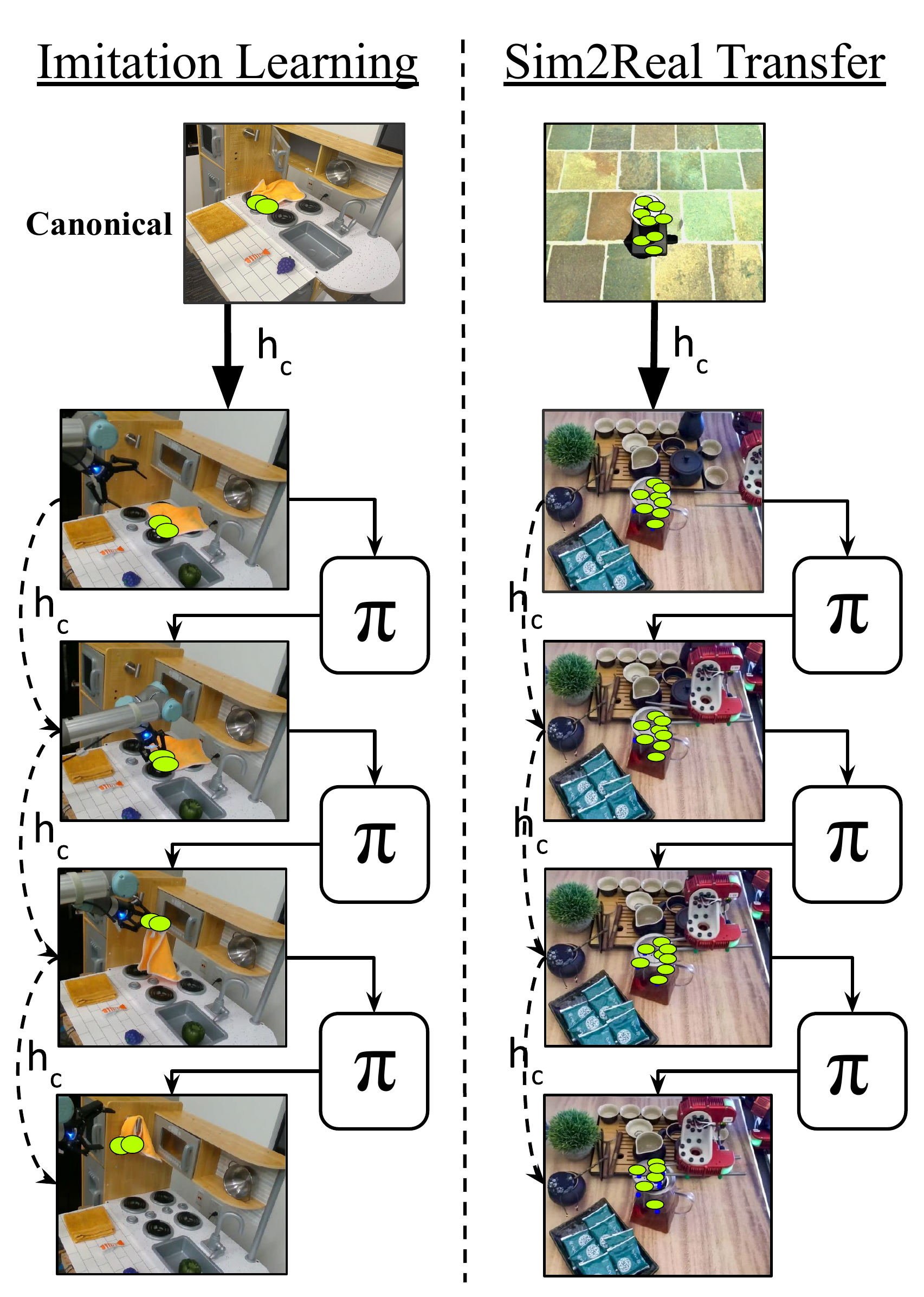}
    \caption{\footnotesize{\textbf{Inference loop with keypoint-based representations.}}}
    \label{fig:inference_kp}
    \vspace{-1.4cm}
\end{wrapfigure}
Assuming the correspondence function \(h_{\mathcal{C}}\) provides confidence scores for each candidate pair, at test time we select the initial‐image points with the highest scores relative to the training canonical keypoints \(\{k_i\}_{i=1}^N\), ensuring accurate matches. These transferred keypoints are then tracked via \(h_{\mathcal{C}}\) and used directly as input to the deployed policy \(\pi_\theta\) without requiring additional masking at test time. Though many implementations of \(h_{\mathcal{C}}\), we adopt the method using diffusion-feature ~\cite{tang23dift,roboabc} for initial set transfer from canonical image due to its strong visual robustness and reliable matching, and employ Cotracker~\cite{cotracker} for subsequent tracking. We provide the pseudocode for applying $\Method$ on a new task in Algorithm \ref{alg:atk} and recommended hyperparameter selection in Table \ref{tab:hp_defaults}.


\section{Experiment Evaluation and Results}

Our experiments aim to answer the following questions: \textbf{(1) Sim-to-real transfer:} How well do the keypoints and policies learned in simulation \emph{transfer} to the real world? \textbf{(2) Imitation robustness:} How well does the proposed keypoint selection and policy learning method work in the imitation learning setting? \textbf{(3) Interpretability and task-relevant features:} Are the chosen keypoints interpretable and relevant to different task objectives in multi-functional environments?


\subsection{Experiment Setup}
 
\subsubsection{Simulation-to-Reality Transfer}

\textbf{Tasks and Challenges.} We consider \textit{three fine manipulation tasks for quantitative analysis}, shown in Fig~\ref{fig:task_in_real}. (1) The \textit{sushi pick-and-place task} requires grasping a piece of sushi in a cluttered environment with distracting objects. (2) The \textit{glasspot tip lifting task} requires precise grasping and lifting of the tip of a glass pot. This task is particularly challenging due to the pot's reflective surface and the small size of the tip. (3) The \textit{clock manipulation task} contains two distinct subtasks: turning the button at the top of the clock or turning the clock hand on its surface, requiring task-specific representations for manipulation of articulated objects. Each task involves the challenges of  tracking difficulty, precision of manipulation, persistence of task-specific focus, and management of variations in scene configuration.

\subsubsection{Robust Imitation Learning}

\begin{figure}[!h]
    \centering
    \includegraphics[width=0.9\linewidth]{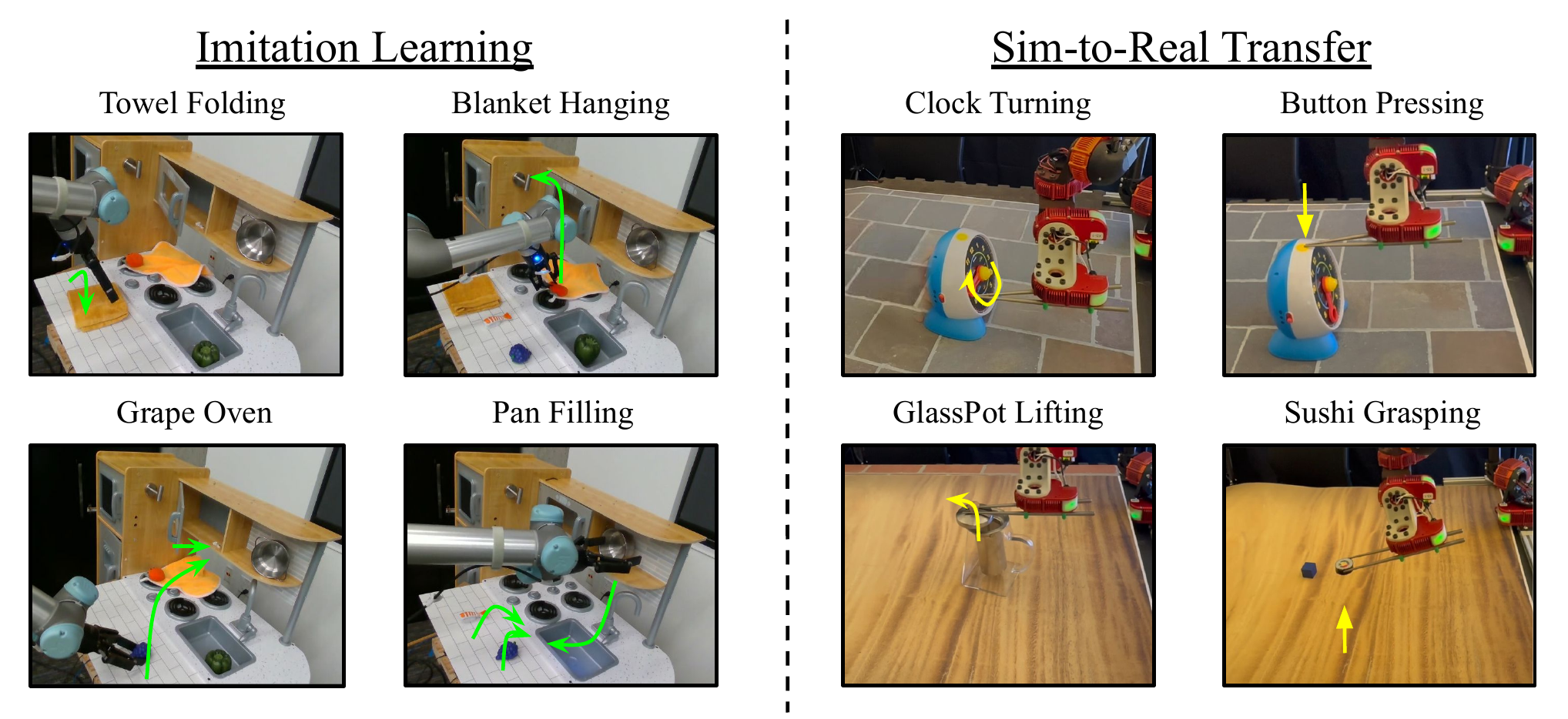}
    \caption{\textbf{Overview of evaluation tasks.} \textit{Left:} Tasks used in an imitation learning setting. \textit{Right:} Tasks used to assess sim-to-real transfer.}
    \label{fig:task_in_real}
\end{figure}

\textbf{Tasks and Challenges.} We consider \textit{four manipulation tasks for quantitative analysis} in a multifunctional kitchen environment as shown in Fig~\ref{fig:task_in_real}. (1) The \textit{grape-oven} task requires grasping a small grape toy and placing it in the microwave. (2) The \textit{blanket hanging} task requires picking up and hanging a deformable object on a hook. (3) The \textit{towel folding} task involves manipulating precise deformable objects without many distinct markers. (4) The \textit{pan filling} task involves transporting and placing a pan on the burner and then placing two different objects, sushi and grapes, into this pan. Notably, each task occurs in the same environment, making it natural to have a focused task-specific representation that is specified to each problem. 



\subsubsection{Baselines and Evaluation} 
\textbf{Baselines.} For both sim-to-real and imitation learning experiments, we compare our approach to two groups of three baselines. 
\emph{(1) Input modality:} 
Policies trained with different input types: \textbf{RGB images}, \textbf{Depth images}, and \textbf{Point clouds}. We use pre-trained visual encoders~\cite{nair22r3m} for RGB and Depth baselines and train pointcloud baselines end-to-end. For detailed training details and architecture, see Appendix \ref{app:baseline}. \emph{(2) Keypoint selection methods:} We consider three more baselines: \textbf{FullSet} uses all sampled keypoints across the image plane; \textbf{Random Select} randomly selects the same number of keypoints as our method; and \textbf{GPTSelect} uses GPT-4 to select the same number of keypoints based on the image and task. 


\textbf{Evaluation.} For each setting, we evaluate each agent on 20 trajectories in the real world, all with varying initial configurations. We illustrate a small range of the randomizations in Fig \ref{fig:concept} and a complete range of randomizations during evaluations in the Appendix. To assess robustness and generalization, we introduce disturbances: \textbf{RP} (random object poses), \textbf{RB} (background texture shuffling), \textbf{RO} (random distractor objects), and \textbf{Light} (altered lighting). For further details on these components, see Appendix.\ref{app:baseline}.

\subsection{Sim-to-Real Transfer of Keypoint-based Policies}

\begin{figure}[H]
    \centering
    \begin{subfigure}[b]{0.49\linewidth}
        \centering
        \includegraphics[width=\linewidth]{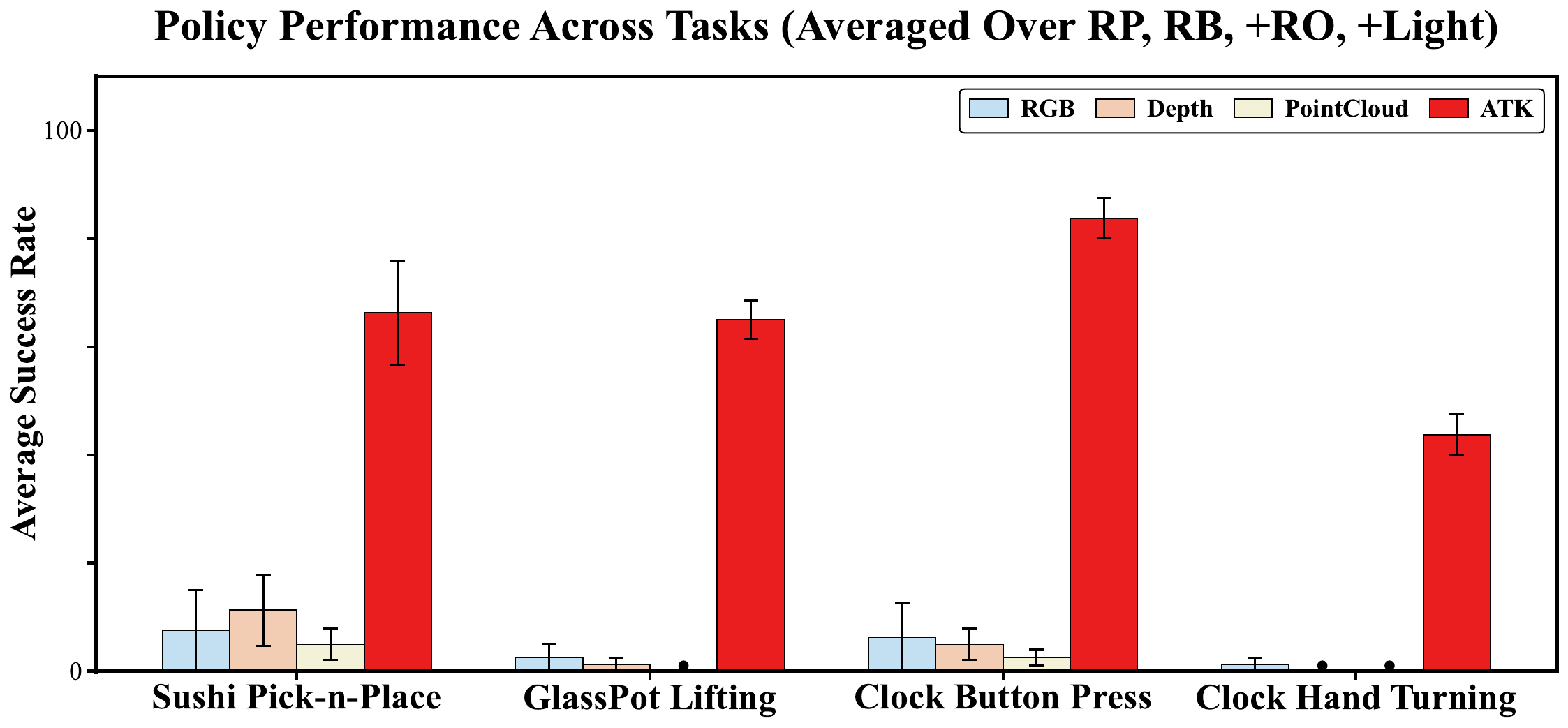}
    \end{subfigure}
    \hfill
    \begin{subfigure}[b]{0.49\linewidth}
        \centering
        \includegraphics[width=\linewidth]{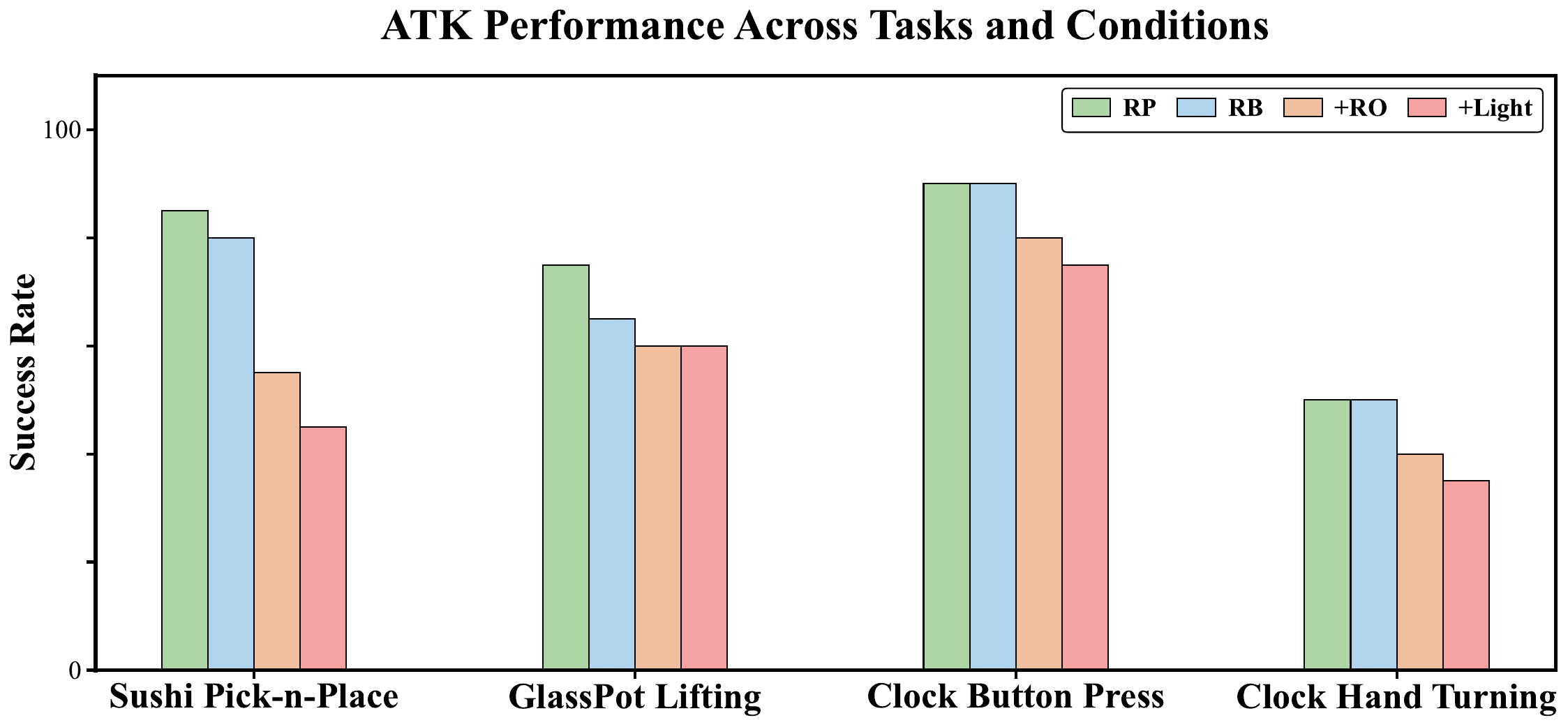}
    \end{subfigure}
     \caption{\textbf{Sim-to-real policy success rates in real world.} \textit{Left:} Aggregated results across real-world evaluation conditions—random pose, background variation, distractor objects, and lighting changes—show that \Method{} outperforms other methods using different input modalities. \textit{Right:}
     \Method{} demonstrates strong robustness under positional variation and various visual perturbations.
}
    \label{fig:sim2real_performance}
\end{figure}

\textbf{$\Method$ transfers from sim-to-real demonstrating visual robustness.} As shown in Fig \ref{fig:sim2real_performance}, keypoint-based policies maintain high success rates in the real world compared to alternative modalities, showcasing strong resilience against randomized object poses or background variations. We provide aggregate performance across different distractors, providing a detailed per-disturbance breakdown in the Appendix. \ref{app:more_res}. Although extreme distractions, such as flashing light or occlusions, can disrupt tracking and decrease performance, \Method  
consistently outperforms RGB, depth, and point-cloud based policies at transfer. The gap is worth noting in tasks involving transparent objects (e.g., glass) and fine-grained manipulation (e.g., clock tasks).
\vspace{-1.8em}  
\subsection{Robust Imitation Learning with Keypoint-based Policies}

\textbf{Policies learned via imitation learning with $\Method$ are resilient to visual disturbances.} We show that policies learned atop the \Method selected representations can perform tasks given significant visual variations. We introduce variations in object positions, backgrounds, distractor objects, and lighting conditions during evaluation. Despite not being trained in these conditions, the learned policies show significantly better transfer performance than other representations (RGB/depth/pointclouds). Moreover, we find that the particular choice of keypoints is crucial for robust transfer performance: FullSet has redundant keypoints that vary significantly, while RandomSelect and GPTSelect often miss important portions of the scene.

\begin{figure}[ht]
    \centering
    \begin{subfigure}[b]{0.49\linewidth}
        \centering
        \includegraphics[width=\linewidth]{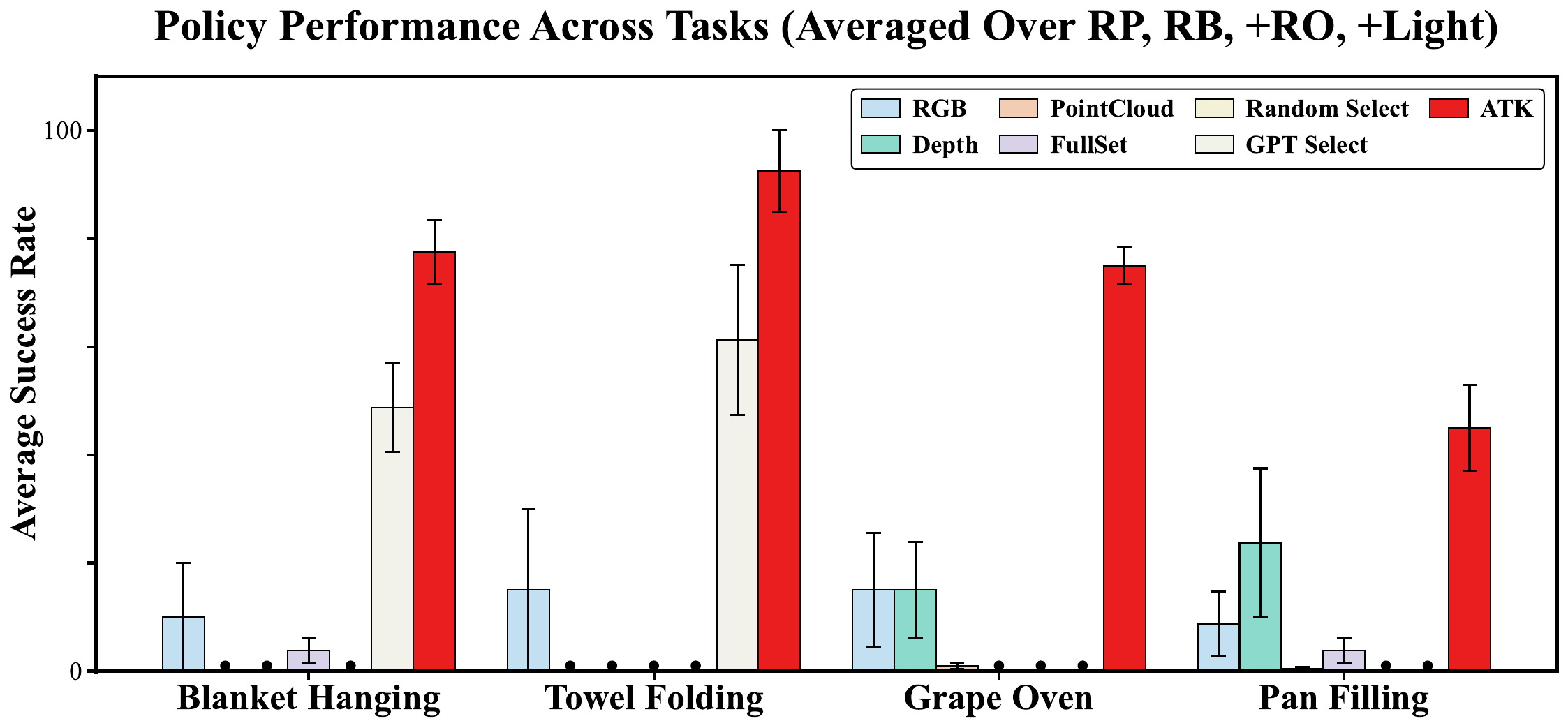}
    \end{subfigure}
    \hfill
    \begin{subfigure}[b]{0.49\linewidth}
        \centering
        \includegraphics[width=\linewidth]{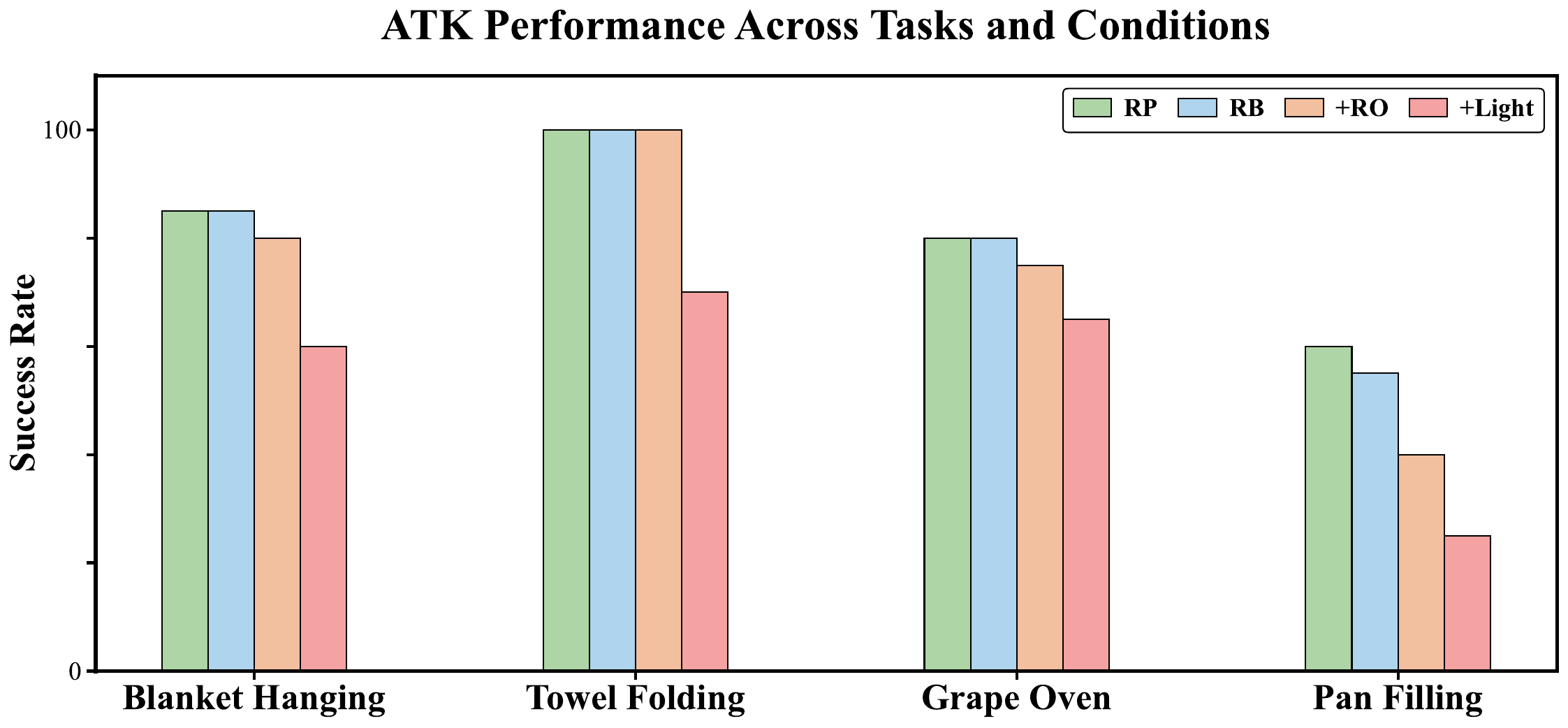}
    \end{subfigure}
    \caption{\textbf{Imitation policy success rates.} \textit{Left:} Aggregated results across diverse evaluation conditions show that \Method{} outperforms other methods based on different input modalities and selection strategies. \textit{Right: }\Method{} demonstrates strong robustness under positional variation and various visual perturbations.}
    \label{fig:imitationperformance}
\end{figure}



\subsection{Qualitative Visualization of Keypoints Learned by $\Method$}

\begin{figure}[!h]
    \centering
    \includegraphics[width=\linewidth]{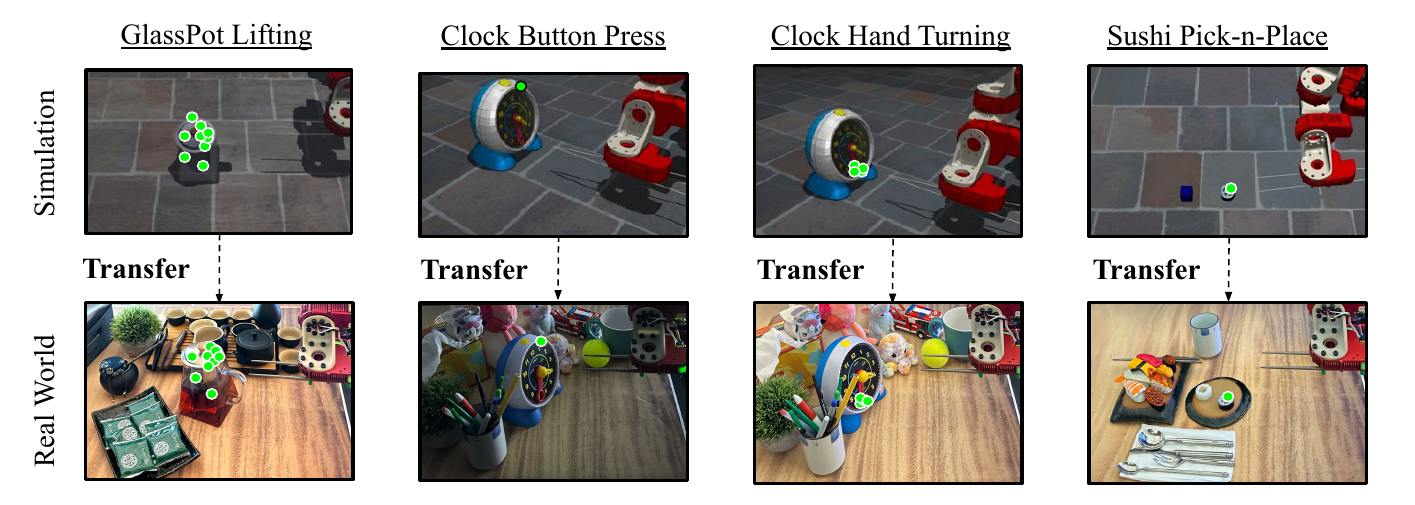}

    \caption{\textbf{Qualitative visualization of task-relevant keypoint selection and transfer.}
Keypoints selected in simulation transfer to real-world scenes across various object positions, backgrounds, distractors, and lighting.}

    \label{fig:qualkeypoints}
    \vspace{-0.1cm}
\end{figure}

\textbf{$\Method$ chooses interpretable and task-relevant keypoints}. In Fig~\ref{fig:qualkeypoints}, we show the keypoints $\Method$ selects, focusing on task-relevant parts. Corresponding visualizations for imitation experiments are done in Fig~\ref{fig:concept}.The chosen keypoints correspond to semantically meaningful elements of the scene. In multifunctional cases (e.g., kitchen, clock), they are specialized to the utility. We see that chosen keypoints transfer accurately from simulation to the real world and from training to testing. The selected keypoints are resilient to visual variations in the scene, including distractors, lighting changes, and background changes.\\ 

\section{Related Work}

\textbf{Visual Representations.} Prior research has explored various visual representation learning approaches for robotics~\cite{laskin20curl,ma23vip,nair22r3m,zhang21bisim,sermanettcn} that use both self-supervised and supervised objectives~\cite{sermanettcn, laskin20curl}. These representations often rely on large-scale pretraining or auxiliary information-theoretic objectives~\cite{ma23vip, zhang21bisim}. Although such representations accelerate policy learning, they often entangle task-irrelevant features and become brittle under distribution shift. In contrast, this work focuses on exploiting privileged expert demonstrations to derive task-driven visual representations suitable for sim-to-real transfer and robust imitation learning.

\textbf{Sim-to-real Transfer.} Bridging the perceptual gap between simulation and the real world remains a significant challenge due to discrepancies in the observation space. Though simulations have become more photorealistic~\cite{morrical2021nvisii}, the direct transfer of policies across domains continues to suffer from performance degradation. Prior work has proposed various methods to mitigate this gap, including domain randomization ~\cite{tobin17rand}, latent representation learning~\cite{nair22r3m, yunlpsim}, unsupervised image translation~\cite{jamessim2sim, ho21retinagan}, depth-based policy~\cite{torne24rialto, chen2023visual} and explicit pose estimation~\cite{handa23dextreme}. Though promising, these methods still face challenges in handling complex, precise tasks, or they rely on task-specific scaffolding (to estimate the pose of a certain object) or restrictive assumptions (e.g., the availability of accurate depth sensors). In this work, we focus on task-driven objectives for visual representations that are intrinsically robust to sim-to-real perturbations.

\textbf{Keypoints as Representations for Learning-based Control.} Keypoints have been utilized as robust state representations for robotic manipulation in several prior works~\cite{wen24anypoint, vecerikrobotap, manuelli19kpam, manuelli20key}. They have been applied in areas including deformable object manipulation~\cite{ma22deformable, sundaresan21knots}, few-shot imitation learning~\cite{vecerikrobotap}, model-based reinforcement learning~\cite{manuelli20key}, and learning from videos~\cite{wen24anypoint, bharadhwaj2024track2act}. However, these approaches often rely on heuristic or manual keypoint selection~\cite{liu2024moka,wen2023anypoint}. Inspired by~\cite{sharma2021generalizing}, our work differs by introducing a task-driven method for automatic keypoint selection. The procedure produces flexible but expressive representations that generalize across rigid, deformable, and transparent objects, and—crucially—are robust to challenging visual disturbances.

\section{Limitations and Conclusion}
We present \Method, a system for automatically selecting task-relevant keypoints, learning keypoint-based policies from these representations, and successfully transferring them to the real-world evaluation scenario. Though promising, the system faces challenges in tracking and optimization. The use of 2D keypoints makes the policy sensitive to camera perspective changes, and off-the-shelf tracking modules may lack robustness for robotic applications with uncommon visual data. Additionally, the method is sensitive to hyperparameters due to the non-smooth nature of the optimization problem, making tuning difficult. Currently, the method trains different models for each individual task. A promising future direction is to incorporate a task-conditional module that can isolate task-specific keypoints and support a unified multi-task selector for diverse manipulation skills.
Nevertheless, our work demonstrates the robustness of keypoint-based policies and provides an effective approach for automatic keypoint selection. Developing more automated and robust techniques to address these challenges would further extend its applicability.



\tcbset{
  gptbox/.style={
    colback=white,
    colframe=accent,
    fonttitle=\bfseries,
    title={\faCommentDots\quad GPT Output},
    boxrule=0.7pt,
    arc=3pt,
    left=6pt,
    right=6pt,
    top=6pt,
    bottom=6pt
  }
}
\acknowledgments{
This work was (partially) funded by grants from the National Science Foundation NRI (\#2132848), DARPA RACER (\#HR0011-21-C-0171), the Office of Naval Research (\#N00014-24-S-B001 and \#2022-016-01 UW), and funding from the Toyota Research Institute through the University Research Program 3.0. We gratefully acknowledge gifts from Amazon, Collaborative Robotics, Cruise, and others.  We would also like to thank Emma Romig, Grant Shogren for their help in setting up the hardware environments for this effort. We thank members of the Washington Embodied Intelligence and Robotics Development Lab and the Personal Robotics Lab for their thoughtful comments and feedback on versions of this draft.}

\clearpage

\newpage
\appendix
\onecolumn

\section{ATK implementation details}
\label{app:Alg}
\subsection{Mask network}
To ensure the selection of the minimal set of $K$ keypoints, we enforce a sparsity penalty on the mask network. We denote the masking distribution over binary masks \(m \in \{0,1\}^N\) for candidate keypoints \(K\) as \(\mathbb{M}_{\phi}(m\mid K),\) and the action distribution conditioned on the masked keypoints as \(\pi_{\theta}\bigl(a \mid \mathbb{M}_{\phi}).\) Because sampling \(m \sim \mathbb{M}_{\phi}(m \mid K)\) is discrete and non-differentiable, we employ the Gumbel–softmax relaxation~\cite{gumbelsoftmax} to enable gradient-based optimization. Concretely, a $N$-dimensional learnable parameters are used as logits for N independent Bernoulli probabilities, with each representing the likelihood of retaining the corresponding keypoint from the input candidate set. We then add Gumbel noise to each logit, divide by a temperature $\tau$, and apply softmax to form a continuous ``soft" mask $y^{Soft}_{M}$. During the forward pass, we take \(
\arg\max_i\,y^{\mathrm{Soft}}_i,\)
to get a hard one-hot selection, but in the backward pass we use the smooth ``soft" mask for gradient calculation. The policy $\pi_{\theta}$ can be instantiated by any policy class model—e.g., a multi-layer perceptron, a Gaussian policy~\cite{rasmussen2003gaussian} or a score-based diffusion model~\cite{chi2023diffusion}. The final objective jointly optimizes the policy \(\pi_{\theta}(a \mid \mathbb{M}_{\phi})\) and the masking distribution \(\mathbb{M}_{\phi}(m \mid K)\), yielding a minimal, task-relevant keypoint representation. 

\subsection{Viewpoint robustness}
\begin{figure}[!h]
    \centering
    \vspace{-0.4cm}
    \includegraphics[width=0.85\linewidth]{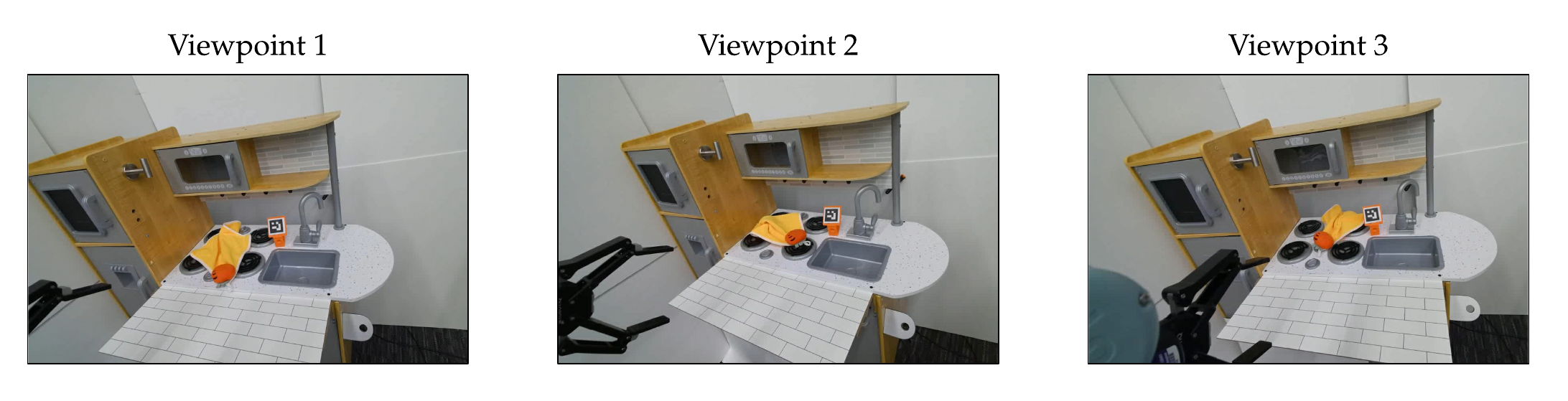}

    \caption{Different camera viewpoint} 
    \label{fig: view}
\end{figure}

“Viewpoint robustness” is another critical metric for gauging how well a policy holds up when the camera’s perspective shifts. It measures the policy’s performance under changed camera viewpoints. In our method, we assume access to both the original and shifted camera intrinsic and extrinsic matrices—a practical assumption given modern computer vision advances(e.g., extrinsic estimation via CtRNet \cite{lu2024ctrnet}). In our experiments, we utilized a fixed ArUco marker's coordinate to get the cameras' extrinsic. We use \( h_{\mathcal{C}} \) to find correspondence 2D keypoints in the new camera view and project them back into the original camera frame where the policy was trained. This reprojection compensates for viewpoint shifts, allowing the policy to operate as if the view had not changed. We show that in our video, after changing the camera to three different angles like Figure \ref{fig: view}, the reprojected keypoints still enable the policy to succeed.
\vspace{-0.5cm}

\subsection{Category-Level Generalization}
\begin{figure}[H]
    \centering
    \includegraphics[width=0.8\linewidth]{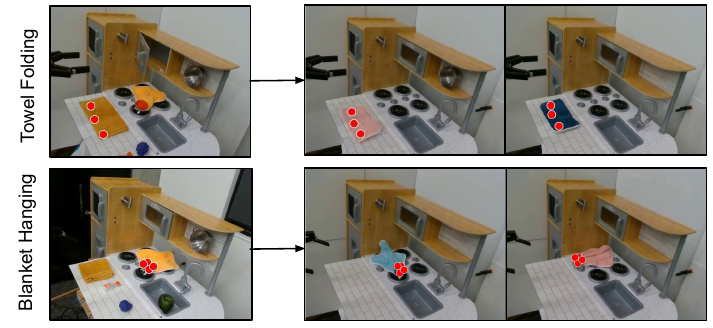}

    \caption{Category Level Generalization} 
    \label{fig: cat}
\end{figure}

To evaluate whether this method can generalize to different objects within the same category, we tested it on two tasks as shown in Figure \ref{fig: cat}. We assume that simulated objects may differ in color, size, or shape from their real-world counterparts; as long as they share the same category, a web-scale pretrained correspondence model can still match them reliably across various visual domain shifts. To measure the robustness of the correspondence matching algorithm, we consider two metrics: (1) Confidence Score, the mean cosine similarity of matched feature vectors, and (2) Mean Distance Error, the average pixel offset between manual annotations and the correspondences predicted by \( h_C \). As shown in the Table \ref{tab:correspondence_results} , the correspondence matching method achieves a high confidence score (0.76--0.82) and a low distance error (6 pixels, < 5\% of the object size), demonstrating accurate sim-to-real keypoint transfer.

\vspace{-0.5cm}

\begin{table}[htbp]
    \centering
    \begin{tabular}{lcccc}
        \toprule
        \textbf{Method} & \textbf{Sushi} & \textbf{Glass} & \textbf{Clock Button} & \textbf{Clock Turning} \\
        \midrule
        Confidence Score & 0.78 & 0.76 & 0.80 & 0.82 \\
        Mean Distance Error (pixels) & 3.24 & 5.21 & 2.79 & 6.75 \\
        \bottomrule
    \end{tabular}
   \vspace{0.2cm}
    \caption{Correspondence Matching Performance Metrics}
    \label{tab:correspondence_results}
\end{table}
\vspace{-0.5cm}
\subsection{High-precision task}
We also evaluate $\Method$ on a \textbf{high-precision manipulation task, i.e., shoe lacing}. For this fine-grained insertion task, the robot needs to insert a shoelace into a shoe eyelet. The task requires high precision since the shoe eyelet has a diameter of 5 mm and the lace's radius is approximately 3.2 mm, leaving only a 1.7 mm tolerance—significantly tighter than the tolerances typically encountered in standard picking or grasping tasks. We tested the robustness of our method under challenging conditions, including varying background textures, random distractors, and changes in lighting. Despite these perturbations, our approach demonstrates high performance.

\begin{figure}[t]
    \centering
    \vspace{-0.4cm}
    \includegraphics[width=.99\linewidth]{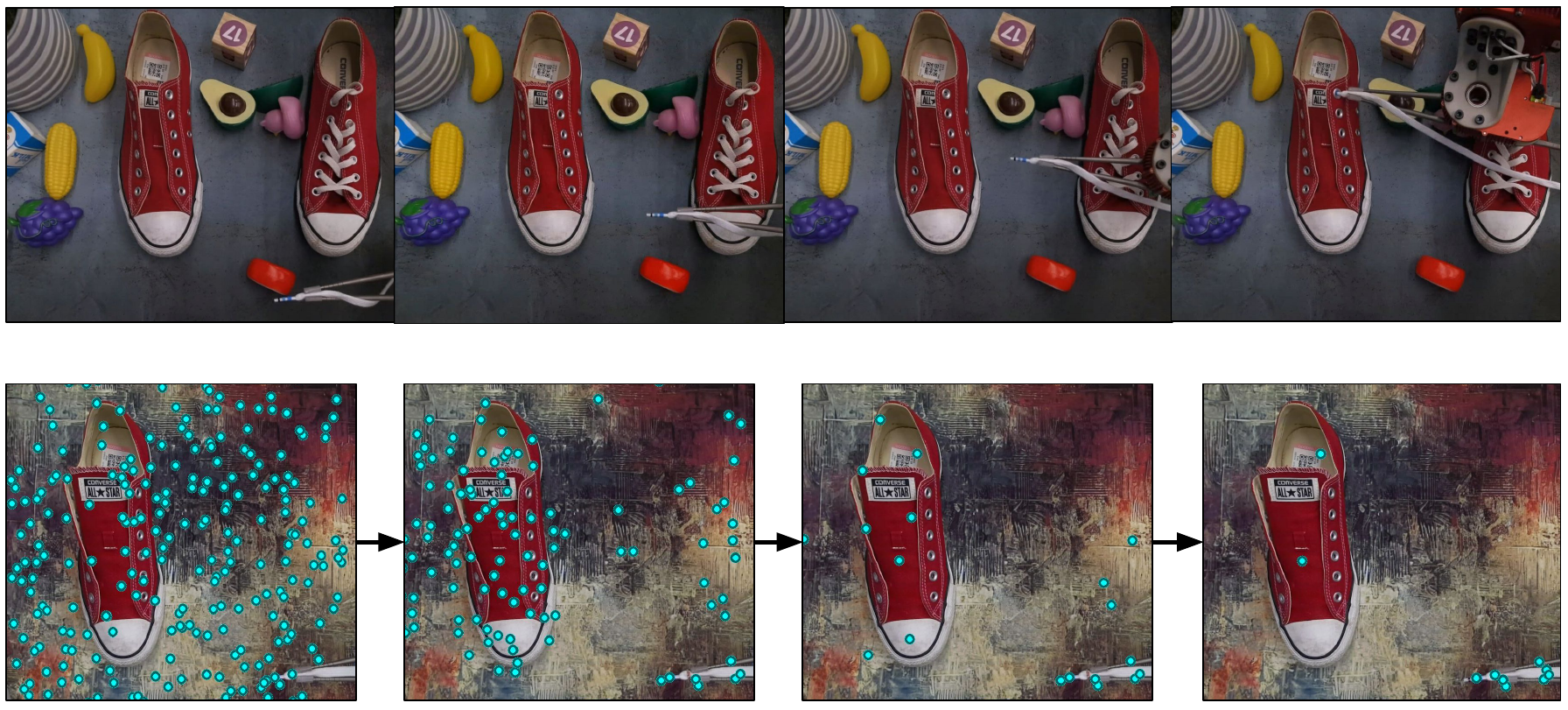}
    \caption{\textbf{The shoe lacing task.} The first row shows a successful rollout of the policy,  performing shoelace insertion with varying backgrounds and distractors. The second row illustrates the keypoint distillation process. }
    \label{fig:shoe_lacing}
    \vspace{-0.5cm}
\end{figure}
\begin{algorithm}[]
\caption{ATK: Automatic Task-Driven Keypoint Selection for Robust Policy Learning}

\label{alg:atk}
\begin{algorithmic}[0]
\Require Expert dataset $\mathcal D=\!\{(o_i,a_i)\}_{i=1}^{M}$,
         correspondence function $h_{\mathcal C}$
\Ensure  Mask model $\mathbb M_{\phi}$, keypoint policy $\pi_{\theta}$

\State \textbf{// Initialize keypoints on canonical frame}
\State Choose canonical image $I_{\text{can}}$ capturing full task context
\State Sample $C$ candidate keypoints $K_{\text{can}}\!=\!\{k_j\}_{j=1}^C$ on $I_{\text{can}}$

\Statex \textbf{// Propagate \& label entire dataset}
\For{$i \gets 0$ \textbf{to} $M-1$}   \Comment{for all trajectories}
  \For{$t \gets 0$ \textbf{to} $T_i - 1$}
    \If{$t = 0$}
      \State $K_{i,0} \gets h_{\mathcal C}\!\bigl(K_{\text{can}},\,o_{i,0},\,I_{\text{can}}\bigr)$
          \Comment{initial alignment to canonical frame}
    \Else
      \State $K_{i,t} \gets h_{\mathcal C}\!\bigl(K_{i,t-1},\,o_{i,t}\bigr)$
          \Comment{track from previous frame}
    \EndIf
  \EndFor
\EndFor
\State Construct $\mathcal D_k=\{(o_{i,t},K_{i,t},a_i)\}$

\Statex \textbf{//Training}
  \While{\textbf{not} converged}    
    \State Sample mini-batch $\mathcal B \subset \mathcal D_k$
    \State Draw mask $m \sim \mathbb M_{\phi}(K)$        \Comment{Gumbel–Softmax}
    \State $\widetilde K \gets m \odot K$                \Comment{retain selected points}
    \State $\mathcal L \gets -\log \pi_{\theta}(a \mid \widetilde K) + \alpha\lVert m\rVert_1$
    \State Update $\phi,\theta$ with Adam to minimise $\mathcal L$
  \EndWhile

\State \Return $(\mathbb M_{\phi},\pi_{\theta})$
\vspace{0.4em}
\Function{Evaluation}{}
    \State \textbf{// Transfer keypoints to test scene}
    \State $K_{0}\gets h_{\mathcal C}\!\bigl(K_{\text{can}},\,I_{\text{can}},\,O_{init}\bigr)$ \Comment{highest-score matches}
    \State Track $K_t=h_{\mathcal C}(K_{t-1},o_t)$ for $t\ge1$
    \State \textbf{return} $\pi_{\theta}(K_t)$ at each step
\EndFunction
\end{algorithmic}
\end{algorithm}
\section{Tasks evaluation procedure}
\subsection{Evaluation metrics}
\begin{figure}[!h]
    \centering
    \vspace{-0.4cm}
    \includegraphics[width=0.8\linewidth]{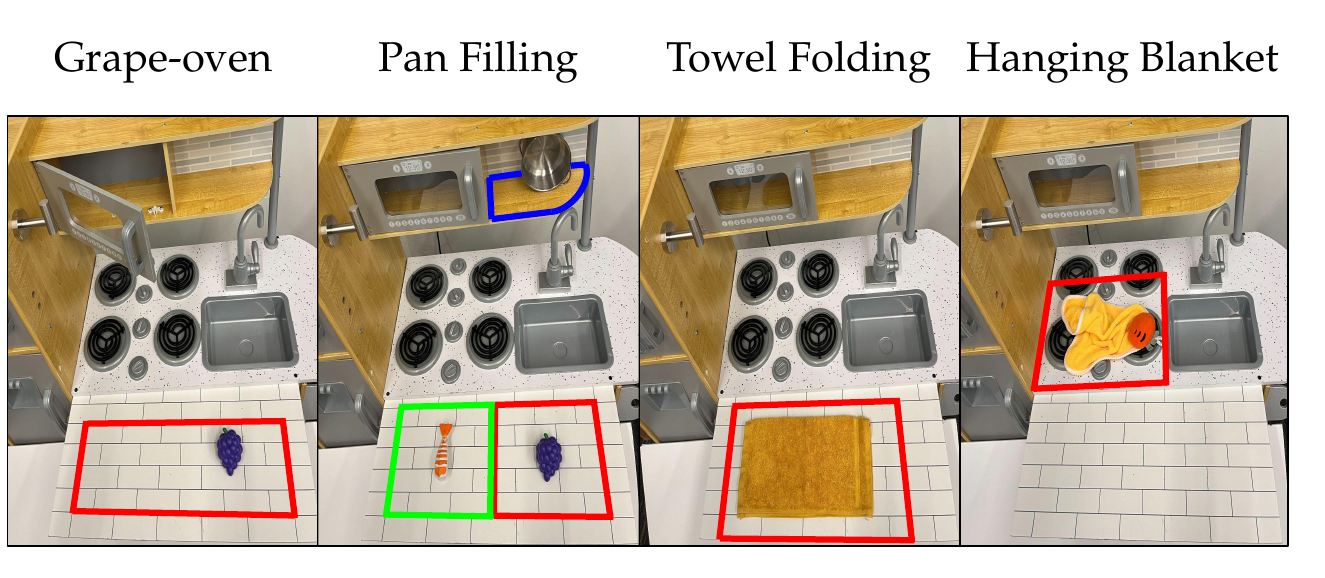}

    \caption{Highlighted regions for different objects during evaluation.} 
    \label{fig: rp}
\end{figure}

\textbf{Random pose (RP):}
As shown in Figure \ref{fig: rp}, we show the distribution of the object's pose during task evaluation.
\\

\begin{figure}[H]
    \centering
    \vspace{-0.4cm}
    \includegraphics[width=0.8\linewidth]{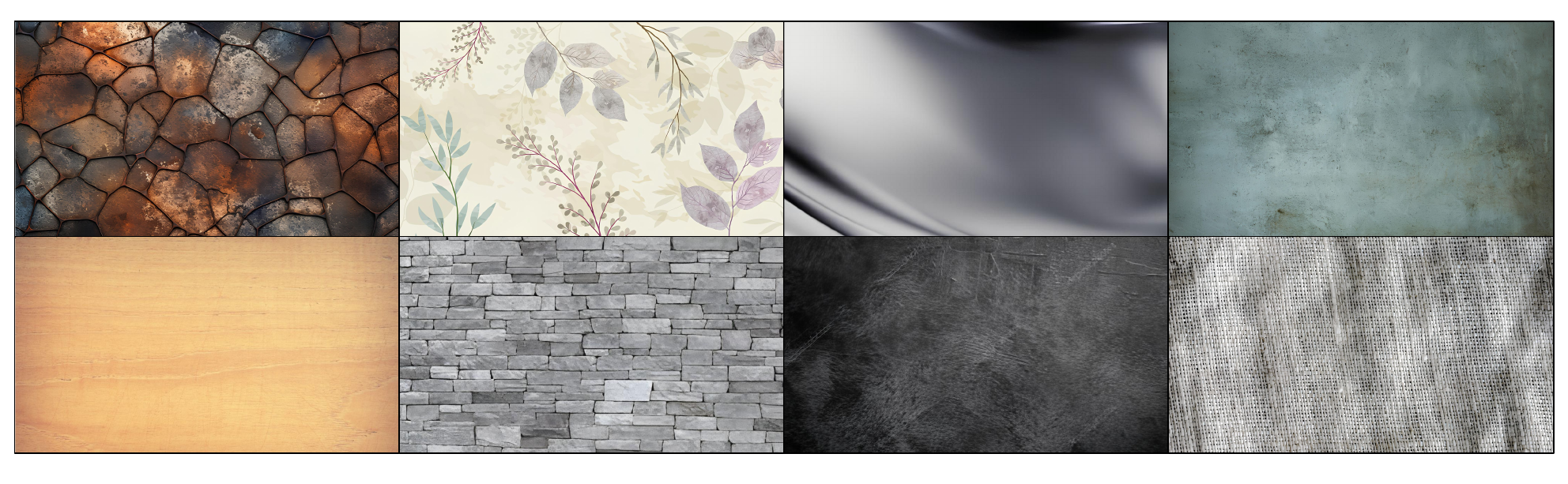} \\

    \caption{Different backgrounds used for evaluation} 
    \label{fig: rb}
\end{figure}

\textbf{Random background (RB):} As shown in Figure \ref{fig: rb}, we show different backgrounds used for evaluation in each task

\begin{figure}[H]
    \centering
    \vspace{-0.4cm}
    \includegraphics[width=0.25\linewidth]{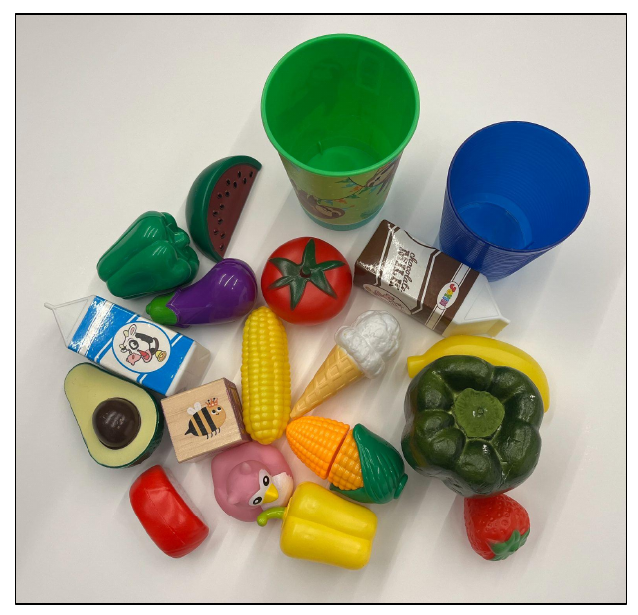}
     \\

    \caption{Distractor Objects used in evaluation} 
    \label{fig: ro}
\end{figure}

\textbf{Random distractor objects (RO):} As shown in Figure \ref{fig: ro}, we show distractor objects used for evaluation in each task.

\begin{figure}[h]
    \centering
    \vspace{-0.4cm}
    \includegraphics[width=0.85\linewidth]{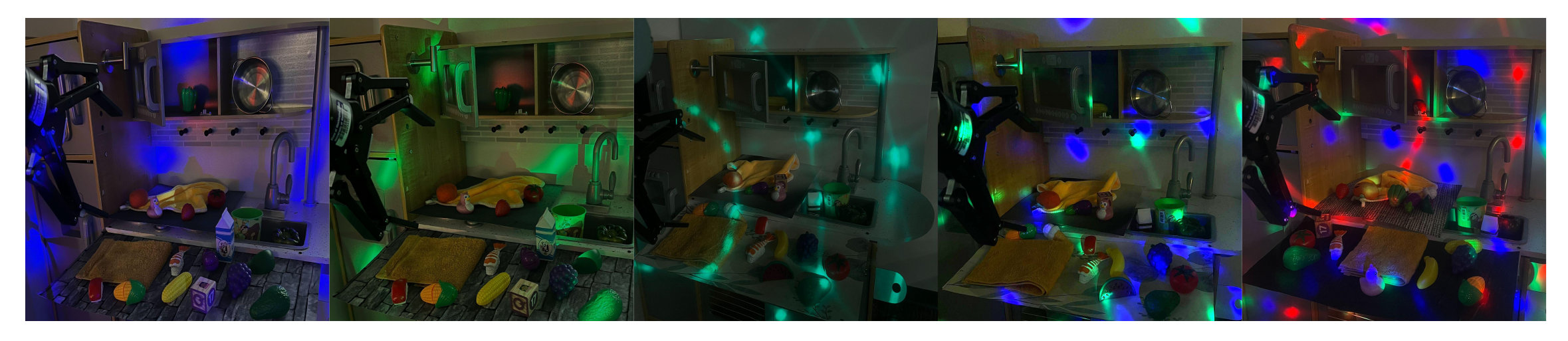}
    \caption{Different Light Conditions used for evaluation} 
    \label{fig: light}
\end{figure}

\textbf{Light:} As shown in Figure \ref{fig: light}, we show the different colored lights used for evaluation in each task

\section{Infrastructural setup}
\label{app:infra}
\textbf{Sim-to-Real Setup:} We create MuJoCo ~\cite{mujoco} simulation environment using an iPhone app, Scaniverse, to scan and import the meshes of real-world objects and add joints for articulated objects. We conduct real-world transfer experiments using a 6-DOF Hebi robot arm equipped with chopsticks, following ~\cite{zhang2023cherry}. For RGB and depth streaming, we employ Azure Kinect RGB-D cameras.

\textbf{Imitation Setup:} Imitation learning is purely tested in the real world. We test these methods on the UR5e robot equipped with a Robotiq 2F-145 gripper, running a joint PD controller. This robot is tasked with manipulating various objects in a miniature kitchen. As previously mentioned, for RGB and depth streaming, we employ Azure Kinect RGB-D cameras. We used 80, 50, 50, 30 number of demonstrations to train policies for pan filling, grape-oven, towel folding, hanging blanket respectively. 
\vspace{-0.5cm}
\section{Baseline implementation details}
\label{app:baseline}

\textbf{ResNet18 Encoder:}
We use a modified ResNet18 architecture as the encoder for depth inputs. The depth data is first resized to 224$\times$224 and duplicated across three channels to match the expected input format of the ResNet18 backbone. Then, we take the ResNet's 512-dimensional feature embedding as a feature map and concatenate it with a 7-dimensional robot state (six joint values and a binary gripper state). The resulting vector is then passed to the diffusion policy \cite{chi2023diffusion} for action prediction.

\textbf{R3M Encoder:}
To handle RGB inputs, we first resize the RGB images from 640$\times$480 resolution to 224$\times$224. Then, we use R3M\cite{r3muniversalvisualrepresentation} to extract a 512-dimensional feature embedding given image inputs and concatenate this embedding with the 7-dimensional joint and gripper state vector to form the input to the diffusion policy \cite{chi2023diffusion}. 

\textbf{DP3 Encoder:}
For encoding 3D point cloud data, we follow the encoder from DP3\cite{dp3}, a CNN-based architecture that utilizes layer normalization. We first project the depth map into 3D and do a spatial crop to get dense point clouds. We then uniformly down-sample $4096$ points as the input for the 3D encoder.

The output 3D feature embedding with size 1024 is concatenated with the same 7-dimensional robot state vector (six joint values and a binary gripper flag) before being used as input to the diffusion policy \cite{chi2023diffusion}. 

\textbf{Baseline Performance Notes:} All baselines (RGB, depth, point-cloud) first used Appendix E.1 encoders with an MLP policy, but performance was weak, so we switched to the official diffusion-policy. With the same 50–60 demos as ATK, each baseline achieved >0\% success on at least one task. However, visual domain shifts and sensor noise especially for depth and pointcloud caused failures, whereas ATK remained robust.

\textbf{Hyperparameter Selection:}  We provide recommended ranges for the core hyperparameters used across our experiments in Table~\ref{tab:hp_defaults}. The default learning rate of $1\times10^{-5}$ is effective across most tasks. Increasing the batch size often improves performance by providing richer contextual information, which is especially beneficial for stable keypoint masking. The sparsity weight $\lambda$ controls the keypoint numbers; for tasks requiring high precision or fine-grained manipulation, a lower sparsity weight is recommended to preserve more expressive and informative keypoints.
\label{app:hp_selection}
\section{Detailed evaluation results}
\label{app:more_res}
\subsection{Sim-to-real performance}
\begin{table}[H]
    \normalsize
    \centering
    \resizebox{\textwidth}{!}
    {
    \begin{tabular}{ccccccccc}
        \toprule
          & \multicolumn{4}{c}{Sushi} & \multicolumn{4}{c}{Glass} \\
        \midrule
        & RP & RB & +RO & + Light & RP & RB & +RO & + Light\\
        \midrule
        RGB & \textcolor{blue}{\textbf{0.453$\pm$0.262}} & 0.076$\pm$0.041 & \textcolor{blue}{\textbf{0.027$\pm$0.020}} & \textcolor{blue}{\textbf{0.010$\pm$0.014}}
        & \textcolor{blue}{\textbf{0.253$\pm$0.154}} & 0.109$\pm$0.098 & \textcolor{blue}{\textbf{0.020$\pm$0.021}} & 0.000$\pm$0.000 \\
        Depth & 0.255$\pm$0.199 &0.255$\pm$0.199 &0.020$\pm$0.021 & \textcolor{blue}{\textbf{0.010$\pm$0.014}}
        & 0.110$\pm$0.001 &\textcolor{blue}{\textbf{0.110$\pm$0.001 }}&0.000$\pm$0.000 & 0.000$\pm$0.000\\
        Pointcloud & 0.277$\pm$0.088 & \textcolor{blue}{\textbf{0.277$\pm$0.088}} &0.020$\pm$0.021 & 0.000$\pm$0.000
        & 0.033$\pm$0.047 &0.033$\pm$0.047 & 0.000$\pm$0.000& 0.000$\pm$0.000\\
        \rowcolor{Gray}
        \Method & \textbf{0.893$\pm$0.073} & \textbf{0.893$\pm$0.073} & \textbf{0.893$\pm$0.073} &\textbf{0.893$\pm$0.073} & \textbf{0.933$\pm$0.034} &  \textbf{0.933$\pm$0.034} &  \textbf{0.933$\pm$0.034}&  \textbf{0.933$\pm$0.034} \\
        \midrule
        & \multicolumn{4}{c}{Clock button} & \multicolumn{4}{c}{Clock turning} \\
        \midrule
        & RP & RB & +RO & + Light & RP & RB & +RO & + Light\\
        \midrule
        RGB & \textcolor{blue}{\textbf{0.456$\pm$0.293}} & 0.046$\pm$0.017 & \textcolor{blue}{\textbf{0.013$\pm$0.019}} & 0.000$\pm$0.000 
        & \textcolor{blue}{\textbf{0.367$\pm$0.205}} & 0.093$\pm$0.020 & \textcolor{blue}{\textbf{0.013$\pm$0.012}} & 0.000$\pm$0.000 \\
        Depth & 0.290$\pm$0.150 & \textcolor{blue}{\textbf{0.290$\pm$0.150}} & 0.000$\pm$0.000&0.000$\pm$0.000 &
        0.256$\pm$0.264 & \textcolor{blue}{\textbf{0.256$\pm$0.264}}&0.000$\pm$0.000 & \textcolor{blue}{\textbf{0.020$\pm$0.021}}\\
        Pointcloud & 0.107$\pm$0.056 &0.107$\pm$0.056 & 0.010$\pm$0.014& 0.000$\pm$0.000&
        0.077$\pm$0.056 &0.077$\pm$0.056 & 0.010$\pm$0.014&0.010$\pm$0.014 \\
        \rowcolor{Gray}
        \Method & \textbf{0.970$\pm$0.024} & \textbf{0.970$\pm$0.024}  & \textbf{0.970$\pm$0.024}  & \textbf{0.970$\pm$0.024}  & \textbf{0.903$\pm$0.028} & \textbf{0.903$\pm$0.028} & \textbf{0.903$\pm$0.028} & \textbf{0.903$\pm$0.028} \\        
        \bottomrule
    \end{tabular}
    }
    \caption{\textbf{Simulator} Policy Success Rates using \emph{different input modalities} over 3 random seeds. Keypoint-based policies are easier to distill in simulator than other baselines with alternative sensor modalities.}
    \label{tab:state}
    \vspace{-.5em}
\end{table}

\begin{table*}[h]
    \normalsize
    \centering
    \resizebox{\textwidth}{!}
    {
    \begin{tabular}{cccccccccccccccccccccccc}
        \toprule
          & \multicolumn{4}{c}{Sushi Pick-n-Place} &  \multicolumn{4}{c}{GlassPot Lift} &  \multicolumn{4}{c}{Clock Button Press} &  \multicolumn{4}{c}{Clock Hand Turning}&  \multicolumn{1}{c}{Total}\\
        \midrule
        & RP & RB & +RO & + Light& RP & RB & +RO & + Light & RP & RB & +RO  & + Light & RP & RB & +RO  & + Light\\
        RGB & \textcolor{blue}{\textbf{0.30}}& 0.00& 0.00& 0.00 &  \textcolor{blue}{\textbf{0.10}}& 0.00& 0.00& 0.00 &  \textcolor{blue}{\textbf{0.25}}& 0.00& 0.00& 0.00 &  \textcolor{blue}{\textbf{0.05}}& 0.00& 0.00& 0.00 & 0.04 \\
        Depth & 0.25& \textcolor{blue}{\textbf{0.20}}& 0.00& 0.00 &  0.05& 0.00& 0.00& 0.00 &  0.10 & \textcolor{blue}{\textbf{0.10}}& 0.00& 0.00 &  0.00& 0.00& 0.00& 0.00 & 0.04\\
        Pointcloud & 0.10& 0.10& 0.00& 0.00 &  0.00& 0.00& 0.00& 0.00 &  0.05& 0.05& 0.00& 0.00 &  0.00& 0.00& 0.00& 0.00 & 0.02\\
        \rowcolor{Gray}
        \Method  & \textbf{0.85} & \textbf{0.80} & \textbf{0.55} & \textbf{0.45} & \textbf{0.75} & \textbf{0.65} &\textbf{0.60} & \textbf{0.60} & \textbf{0.90} & \textbf{0.90} & \textbf{0.80} & \textbf{0.75} & \textbf{0.50} & \textbf{0.50} & \textbf{0.40} & \textbf{0.35} & \textbf{0.64}\\
        \bottomrule
    \end{tabular}
    }
    \caption{\textbf{Real-world} Policy Success Rates. Varying conditions including RP (random pose), RB (background), RO (distractor object), Light. $\Method$ consistently outperforms baseline methods using alternative modalities in sim-to-real transfer. }
    \label{tab:real}
    \vspace{-.5em}
\end{table*}
\begin{table}[H]
    \normalsize
    \centering
    \resizebox{\textwidth}{!}
    {
    \begin{tabular}{ccccccccc}
        \toprule
          & \multicolumn{4}{c}{Sushi} & \multicolumn{4}{c}{Glass} \\
        \midrule
        & RP & RB & +RO & + Light & RP & RB & +RO & + Light\\
        \midrule
        FullSet & 0.122$\pm$0.057 & 0.053$\pm$0.036 & 0.010$\pm$0.008 & 0.013$\pm$0.012 
        & \textcolor{blue}{\textbf{0.311$\pm$0.150}} & \textcolor{blue}{\textbf{0.069$\pm$0.056}} & 0.013$\pm$0.012 & \textcolor{blue}{\textbf{0.013$\pm$0.012}} \\
        RandomSelect & \textcolor{blue}{\textbf{0.337$\pm$0.315}} & \textcolor{blue}{\textbf{0.246$\pm$0.360}} & \textcolor{blue}{\textbf{0.233$\pm$0.370}}& \textcolor{blue}{\textbf{0.226$\pm$0.375}}
        & 0.120$\pm$0.082 & 0.031$\pm$0.044&\textcolor{blue}{\textbf{0.116$\pm$0.151}} &0.006$\pm$0.009 \\
        GPTSelect & 0.032$\pm$0.009 & 0.020$\pm$0.008 &0.013$\pm$0.005 & 0.006$\pm$ 0.004
        & 0.133$\pm$0.188 & 0.020$\pm$0.028& 0.010$\pm$0.014& 0.010$\pm$0.014\\
        \rowcolor{Gray}
        \Method & \textbf{0.893$\pm$0.073} & \textbf{0.893$\pm$0.073} & \textbf{0.893$\pm$0.073} &\textbf{0.893$\pm$0.073} & \textbf{0.933$\pm$0.034} &  \textbf{0.933$\pm$0.034} &  \textbf{0.933$\pm$0.034} &  \textbf{0.933$\pm$0.034} \\
        \midrule
        & \multicolumn{4}{c}{Clock button} & \multicolumn{4}{c}{Clock turning} \\
        \midrule
        & RP & RB & +RO & + Light & RP & RB & +RO & + Light\\
        \midrule
        FullSet & 0.474$\pm$0.317 & 0.126$\pm$0.090 & 0.026$\pm$0.030 & 0.020$\pm$0.016 
        & 0.253$\pm$0.183 & 0.083$\pm$0.880 & 0.010$\pm$0.014 & 0.010$\pm$0.014 \\
        RandomSelect & 0.107$\pm$0.030 & 0.080$\pm$0.045 &0.036$\pm$0.032 &0.026$\pm$0.020 &
        \textcolor{blue}{\textbf{0.253$\pm$0.166}} &0.076$\pm$0.088 & 0.000$\pm$0.000& 0.000$\pm$0.000\\
        GPTSelect & \textcolor{blue}{\textbf{0.913$\pm$0.041}} &\textcolor{blue}{\textbf{0.913$\pm$0.041}} & \textcolor{blue}{\textbf{0.913$\pm$0.041}}& \textcolor{blue}{\textbf{0.913$\pm$0.041}}&
        0.065$\pm$0.053 &\textcolor{blue}{\textbf{0.146$\pm$0.179}} & \textcolor{blue}{\textbf{0.077$\pm$0.088}}& \textcolor{blue}{\textbf{0.020$\pm$0.028}}\\
        \rowcolor{Gray}
        \Method & \textbf{0.970$\pm$0.024} & \textbf{0.970$\pm$0.024}  & \textbf{0.970$\pm$0.024}  & \textbf{0.970$\pm$0.024}  & \textbf{0.903$\pm$0.028} & \textbf{0.903$\pm$0.028} & \textbf{0.903$\pm$0.028} & \textbf{0.903$\pm$0.028} \\
        \bottomrule
    \end{tabular}
    }
   \vspace{0.2cm}
    
    \caption{\textbf{Simulator} Policy Success rate using \textbf{different keypoint selection methods} over 3 random seeds. $\Method$ consistently outperforms alternative keypoint selection methods using random sampling or ChatGPT selection. }
    \label{tab:select}
    \vspace{-.5em}
\end{table}

\subsection{Imitation learning performance}
\begin{table}[H]
    \normalsize
    \centering
    \resizebox{\textwidth}{!}
    {
    \begin{tabular}{cccccccccccccccccccccccc}
        \toprule
          & \multicolumn{4}{c}{Towel Hanging} &  \multicolumn{4}{c}{Towel Folding} &  \multicolumn{4}{c}{Grape Oven} &  \multicolumn{4}{c}{Pan Filling}&  \\
        \midrule
        & RP & RB & +RO & + Light& RP & RB & +RO & + Light & RP & RB & +RO  & + Light & RP & RB & +RO  & + Light\\
        RGB & \textcolor{blue}{\textbf{0.40}}& 0.00& 0.00& 0.00 &  \textcolor{blue}{\textbf{0.60}}& 0.00& 0.00& 0.00 &  \textcolor{blue}{\textbf{0.45}}& 0.15& 0.00& 0.00 &  \textcolor{blue}{\textbf{0.25}}& 0.10& 0.00& 0.00  \\
        Depth & 0.00 & 0.00 & 0.00 & 0.00 &  0.00& 0.00& 0.00 & 0.00 &  0.35 & \textbf{0.25}& 0.00& 0.00 &  \textcolor{blue}{0.50} & 0.45& 0.00& 0.00 \\
        Pointcloud & 0.00& 0.00& 0.00& 0.00 &  0.00& 0.00& 0.00& 0.00 &  0.02& 0.02& 0.00& 0.00 &  0.01& 0.01& 0.00& 0.00 \\
        \rowcolor{Gray}
        \Method  & \textbf{0.85} & \textbf{0.85} & \textbf{0.80} & \textbf{0.60} & \textbf{1.00} & \textbf{1.00} &\textbf{1.00} & \textbf{0.70} & \textbf{0.80} & \textbf{0.80} & \textbf{0.75} & \textbf{0.65} & \textbf{0.60} & \textbf{0.55} & \textbf{0.40} & \textbf{0.25} \\
        \bottomrule
    \end{tabular}
    }
        \vspace{0.2cm}
    \caption{\textbf{Real World Imitation Learning} Success Rates. Varying conditions including RP (random pose), RB (background), RO (distractor object), Light. ATK consistently outperforms baseline methods using alternative modalities.}
    \label{tab:real}
    \vspace{-.5em}
\end{table}
\begin{table}[H]
    \normalsize
    \centering
    \resizebox{\textwidth}{!}
    {
    \begin{tabular}{ccccccccccccccccc}
        \toprule
          & \multicolumn{4}{c}{Towel Hanging} & \multicolumn{4}{c}{Towel Folding} & \multicolumn{4}{c}{Grape Oven} & \multicolumn{4}{c}{Pan Filling} \\
        \midrule
        & RP & RB & +RO & + Light & RP & RB & +RO & + Light & RP & RB & +RO & + Light & RP & RB & +RO & + Light\\
        \midrule
        FullSet & 0.10 & 0.05 & 0.00 & 0.00 
        & 0.00 & 0.00 & 0.00 & 0.00 & 0.00 & 0.00 & 0.00 & 0.00 & 0.10 & 0.05 & 0.00 & 0.00\\
        RandomSelect & 0.00 & 0.00 & 0.00 & 0.00 & 0.00 & 0.00 & 0.00 & 0.00  & 0.00 & 0.00 & 0.00 & 0.00 & 0.00 & 0.00 & 0.00 & 0.00\\
        GPTSelect & 0.60 & 0.60 & 0.50 & 0.25 & 1.00 & 0.60 & 0.50 & 0.35 & 0.00 & 0.00 & 0.00 & 0.00 & 0.00 & 0.00 & 0.00 & 0.00\\
        \rowcolor{Gray}
        \Method & \textbf{0.85} & \textbf{0.85} & \textbf{0.80} &\textbf{0.60}  &  \textbf{1.00} &  \textbf{1.00} &  \textbf{1.00} & \textbf{0.70} & \textbf{0.80} & \textbf{0.80} & \textbf{0.75} &\textbf{0.65} & \textbf{0.60} &  \textbf{0.55} &  \textbf{0.40} &  \textbf{0.25} \\
      
        \bottomrule
    \end{tabular}
    }
    
    \caption{\textbf{Real World Imitation Learning} Success rate using \textbf{different keypoint selection methods}. $\Method$ consistently outperforms alternative keypoint selection methods using random sampling or ChatGPT selection. }
    \label{tab:select}
    \vspace{-.5em}
\end{table}

\begin{table}[H]
    \caption{Core hyperparameters and their recommended ranges.}
    \label{tab:hp_defaults}
    \centering
    \begin{tabular}{@{}lcc@{}}
        \toprule
        \textbf{Parameter} & \textbf{Default} & \textbf{Typical Range} \\
        \midrule
        Learning rate & $1\times10^{-5}$ & $1\times10^{-5}$–$1\times10^{-4}$ \\
        Batch size & $4096$ & $2048$–$4096$ \\
        Sparsity Weight $\lambda$ & $0.002$ & $0.001$–$0.005$ \\
        \bottomrule
    \end{tabular}
\end{table}

\subsection{GPT selection details}
\begin{tcolorbox}[colback=lightgray,colframe=darkgray,title=Prompt for keypoint selection using GPT for Grape-oven task ]
\textbf{Consider the image:}\\
\begin{center}
    \centering
    \includegraphics[width=0.45\textwidth]{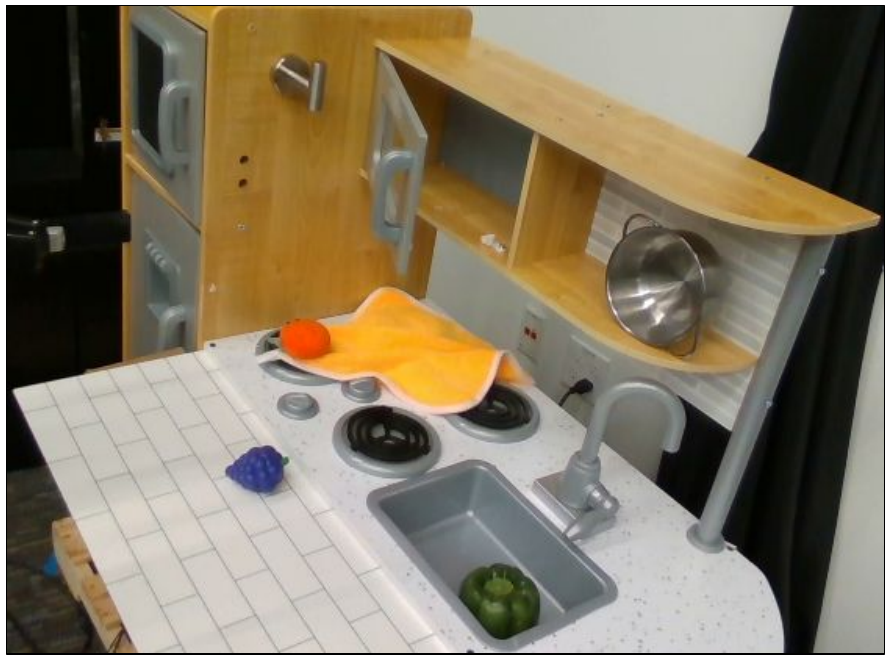 }
\end{center}
\textbf{Task Description:}\\
You are training a robotic arm policy to perform a task that involves:
\begin{enumerate}
    \item  Identifying and picking up a grape object in an image.
    \item Placing the object inside an open gated oven.
    \item Closing the oven door.
\end{enumerate}

\textbf{Keypoint Requirements:}\\
The robot’s control policy uses relevant visual keypoints as input to solve the above task. Your objective is to:
\begin{itemize}
    \item Identify and output 4 critical keypoints in the provided image that will help the policy learn this task.
    \item Draw your output keypoints on the above given image
\end{itemize}

\end{tcolorbox}
\begin{tcolorbox}[colback=lightgray,colframe=darkgreen,title=GPT Output ]
\begin{center}
    \centering
    \includegraphics[width=0.45\textwidth]{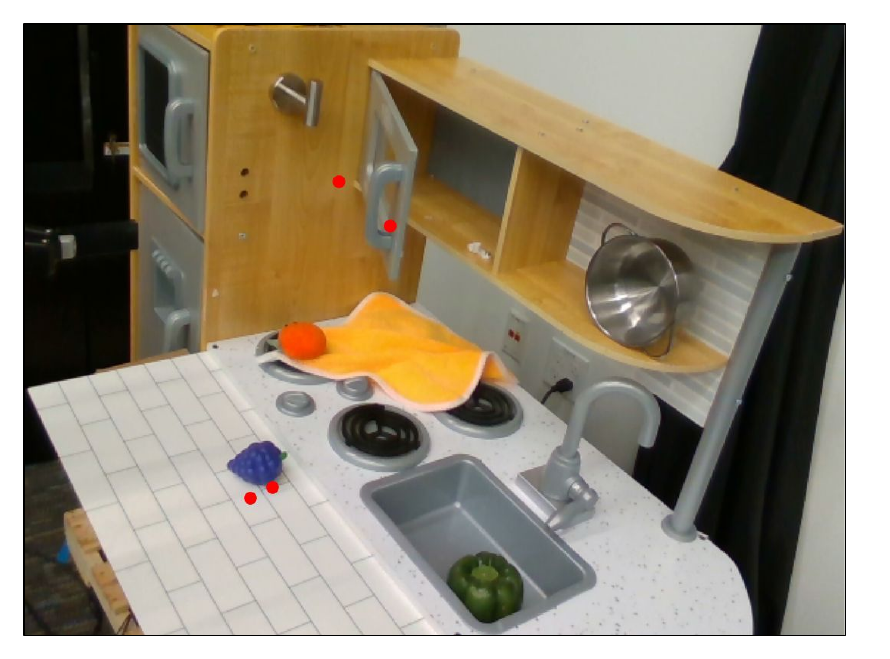 }
\end{center}
Here are the 2D pixel coordinates for the selected keypoints:\\\\
On the Grape Object
\begin{itemize}
    \item Left side of the object: \textbf{(210, 375)}
    \item Right/top side of the object: \textbf{(225, 368)}
\end{itemize}
On the Oven (Above the Stove)
\begin{itemize}
    \item Inside the oven tray (for placing the object): \textbf{(305, 190)}
    \item Oven handle (for closing the door): \textbf{(270, 160)}
\end{itemize}
\end{tcolorbox}

\begin{tcolorbox}[colback=lightgray,colframe=darkgray,title=Prompt for keypoint selection using GPT for Pan-filling task]
\textbf{Consider the image:}\\
\begin{center}
    \centering
    \includegraphics[width=0.4\textwidth]{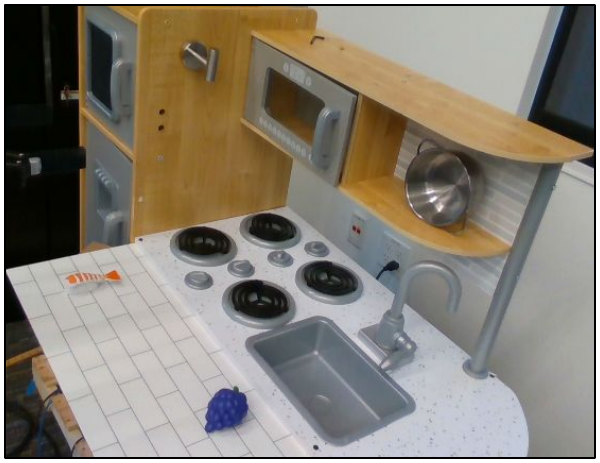 }
\end{center}

\textbf{Task Description:}\\
You are training a robotic arm policy to perform a task that involves:
\begin{enumerate}
    \item Identifying and picking up a pan placed on a shelf above the sink in the above image.
    \item Placing the pan on the right bottom gas burner.
    \item Picking up sushi, placing it inside the pan, and finally picking up a grape and placing it inside the pan.
\end{enumerate}

\textbf{Keypoint Requirements:}\\
The robot’s control policy uses relevant visual keypoints as input to solve the above task. Your objective is to:
\begin{itemize}
    \item Identify and output 6 critical 2D keypoints in the provided image that will help the policy learn this task.
    \item Draw your output keypoints on the above-given image
\end{itemize}

\end{tcolorbox}
\begin{tcolorbox}[colback=lightgray,colframe=darkgreen,title=GPT Output ]
\begin{center}
    \centering
    \includegraphics[width=0.4\textwidth]{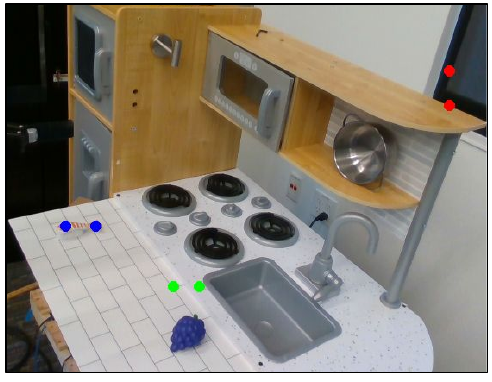 }
\end{center}
Here are the 2D pixel coordinates for the selected keypoints:
Sushi Keypoints:
\begin{itemize}
    \item sushi\_1: (105, 230)
    \item sushi\_2: (125, 240)
\end{itemize}

Grape Keypoints:
\begin{itemize}
    \item grape\_1: (215, 350)
    \item grape\_2: (225, 370)
\end{itemize}

Pan Keypoints:
\begin{itemize}
    \item pan\_1: (495, 160)
    \item pan\_2: (525, 180)
\end{itemize}
\vspace{0.8em}

\end{tcolorbox}
\begin{tcolorbox}[colback=lightgray,colframe=darkgray,title=Prompt for keypoint selection using GPT for Towel Folding task]
\textbf{Consider the image:}\\
\begin{center}
    \centering
    \includegraphics[width=0.45\textwidth]{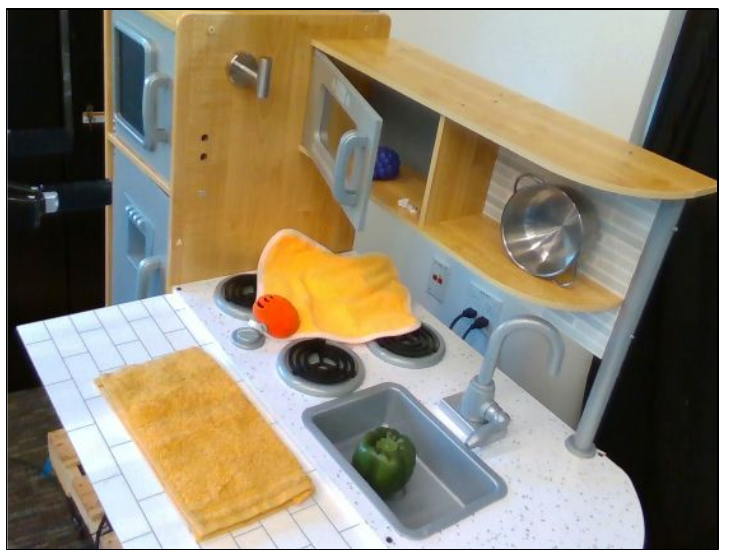 }
\end{center}
\textbf{Task Description:}
You are training a robotic arm policy to perform a task that involves identifying and folding a towel lying flat on the kitchen counter:

\textbf{Keypoint Requirements:}\\
The robot’s control policy uses relevant visual keypoints as input to solve the above task. Your objective is to:
\begin{itemize}
    \item Identify and output 6 critical 2D keypoints in the provided image that will help the policy learn this task.
    \item Draw your output keypoints on the above given image
\end{itemize}

\end{tcolorbox}
\begin{tcolorbox}[colback=lightgray,colframe=darkgreen,title=GPT Output ]
\begin{center}
    \centering
    \includegraphics[width=0.45\textwidth]{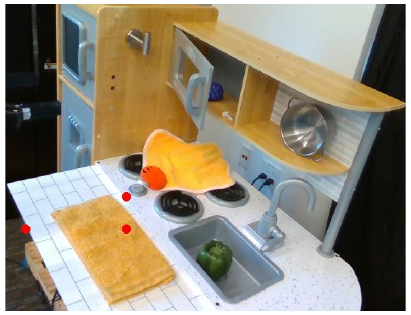 }
\end{center}
Here are the 2D pixel coordinates for the selected keypoints:

\vspace{0.8em}
\begin{itemize}
    \item Left Edge: (85, 365)
    \item Right Edge: (225, 365)
    \item Top Edge: (225, 320)
\end{itemize}
\end{tcolorbox}

\begin{tcolorbox}[colback=lightgray,colframe=darkgray,title=Prompt for keypoint selection using GPT for Hanging Blanket task]
\textbf{Consider the image:}\\
\begin{center}
    \centering
    \includegraphics[width=0.45\textwidth]{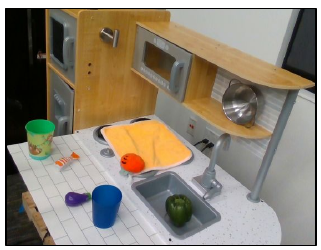 }
\end{center}
\textbf{Task Description:}
You are training a robotic arm policy to perform a task that involves identifying and picking up a blanket lying flat on the gas stove and hang it on a hook attached on a side wall:

\textbf{Keypoint Requirements:}\\
The robot’s control policy uses relevant visual keypoints as input to solve the above task. Your objective is to:
\begin{itemize}
    \item Identify and output 4 critical 2D keypoints in the provided image that will help the policy learn this task.
    \item Draw your output keypoints on the above given image
\end{itemize}

\end{tcolorbox}
\begin{tcolorbox}[colback=lightgray,colframe=darkgreen,title=GPT Output ]
\begin{center}
    \centering
    \includegraphics[width=0.45\textwidth]{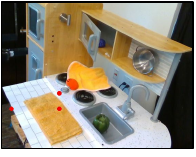 }
\end{center}
Here are the 2D pixel coordinates for the selected keypoints:

\vspace{0.8em}
\begin{itemize}
    \item Grasp Ball (Pickup): (330, 300)
    \item Towel Top Edge: (280, 280)
    \item Towel Center: (300, 320)
    \item Towel Side Edge: (260, 340)
\end{itemize}
\end{tcolorbox}

\label{app:gpt}
\subsection{Additional KeyPoint Selection Baselines}
\begin{figure}[htbp]
    \centering
    \vspace{-0.3cm} 
    \includegraphics[width=0.85\linewidth]{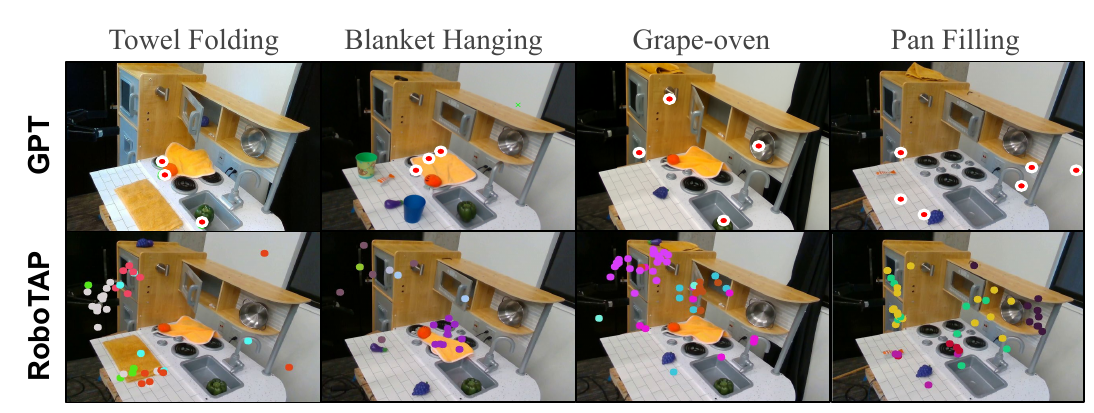}
    \vspace{-0.2cm} 
    \caption{First row: GPT results when prompted with the initial set of candidate keypoints. Second row: RoboTAP selection results. Different colors denote distinct clusters of keypoints remaining after selection.} 
    \label{fig: gpt_robotap}
    \vspace{-0.4cm} 
\end{figure}
To make fairer comparison, we provide GPT with the same set of candidate keypoints as used for the initial set of keypoints on the canonical image from \Method. As illustrated in the figure \ref{fig: gpt_robotap} , the GPT-Select approach occasionally suffers from spatial mismatches (incorrectly locating the relevant point in the image) or language mismatches (incorrectly interpreting the textual prompt), which leads to erroneous keypoint selections. We also evaluated Robotap\cite{robotap} keypoint's selector on our tasks. As shown in the above figure, its design for ego-centric images prevents it from filtering out irrelevant points effectively in our setup.

\end{document}